\documentclass[conference]{IEEEtran}  

\IEEEoverridecommandlockouts                              

\usepackage{enumitem}
\usepackage{graphics} 
\usepackage{amsmath} 
\usepackage{amssymb}  
\usepackage{todonotes}
\usepackage{comment}
\usepackage{diagbox}
\usepackage{makecell}
\usepackage{booktabs}
\usepackage{comment}
\usepackage{float}
\usepackage{mathtools}
\usepackage[linesnumbered,ruled,vlined]{algorithm2e}
\usepackage[noend]{algpseudocode}
\usepackage{varwidth}
\usepackage{ulem}
\usepackage{soul}
\usepackage{cite}
\usepackage{multirow}
\usepackage{tabu}
\usepackage{hyperref}
\usepackage[flushleft]{threeparttable}
\usepackage{capt-of}
\usepackage{amsthm}

\newtheorem{theorem}{Theorem}
\newtheorem{lemma}[theorem]{Lemma}
\newtheorem{example}{Example}

\def\quad{\hskip0.5em\relax}

\DeclareMathOperator*{\minimize}{minimize}

\title{\LARGE \bf
Accelerating Signal-Temporal-Logic-Based Task and Motion Planning of Bipedal Navigation using Benders Decomposition
}

\author{
Jiming Ren$^{*}$, Xuan Lin$^{*}$, Roman Mineyev, Karen M. Feigh, Samuel Coogan and Ye Zhao
\thanks{}
\thanks{The authors are with the Institute for Robotics and Intelligent Machines, Georgia Institute of Technology, Atlanta, GA 30332, USA, {\tt\small \{jren313, xlin373\}@gatech.edu}}
\thanks{*equally contributed}
\thanks{This work was supported in part by the Lockheed Martin Corporation; the Office of Naval Research (ONR) under Grant N000142312223; National Science Foundation (NSF) under Grant IIS-1924978, Grant CMMI-2144309, and Grant FRR-2328254.}
}

\begin{document}
\bstctlcite{IEEEexample:BSTcontrol} 

\maketitle
\thispagestyle{empty}
\pagestyle{empty}

\begin{abstract}
Task and motion planning under Signal Temporal Logic constraints is known to be NP-hard. A common class of approaches formulates these hybrid problems, which involve discrete task scheduling and continuous motion planning, as mixed-integer programs (MIP). However, in applications for bipedal locomotion, introduction of non-convex constraints such as kinematic reachability and footstep rotation exacerbates the computational complexity of MIPs. In this work, we present a method based on Benders Decomposition to address scenarios where solving the entire monolithic optimization problem is prohibitively intractable. Benders Decomposition proposes an iterative cutting-plane technique that partitions the problem into a master problem to prototype a plan that meets the task specification, and a series of subproblems for kinematics and dynamics feasibility checks. Our experiments demonstrate that this method achieves faster planning compared to alternative algorithms for solving the resulting optimization program with nonlinear constraints. A project website can be found at \url{http://bipedal-stl.github.io/}.

\end{abstract}


\textbf{\small{\textit{Note to Practitioners}---}}
\small{\textbf{Bipedal robots are increasingly demanded in warehouses and factories for complex automation tasks such as stacking, delivering, and interacting with other robots under strict time and safety constraints. However, planning such operations under formal language instructions such as Signal Temporal Logic (STL) specifications often results in large-scale mixed-integer programs that are impractical to be solved in a timely manner. This paper introduces an accelerated task and motion planning (TAMP) approach via Benders Decomposition that splits the task into a high-level scheduling problem and lower-level motion feasibility checks, allowing practitioners to find feasible and optimal task and motion plans far more efficiently. Compared to conventional monolithic solvers or alternative decomposition methods, our approach can generate solutions more than twenty times faster while rigorously satisfying kinematic and dynamic constraints. Benchmark scenarios, including factory delivery and warehouse logistics, demonstrate how our method handles realistic automation scenarios involving long planning horizons and complicated task specifications. This makes it practical to deploy bipedal robots for logistics missions that would otherwise be computationally intractable. Practitioners can apply this framework to their own robots and STL tasks to reduce planning time and guarantee safe, feasible motion plans, thereby enhancing automation productivity. While the high-level scheduling problem still dominates runtime, this work shows a clear path toward scalable, reliable TAMP for bipedal robots in industrial environments.}}

\textbf{\small{\textit{Index Terms}---}}
\small{\textbf{Task and Motion Planning, Formal Methods, Temporal Logic, Bipedal Locomotion, Mixed-Interger Program, Benders Decomposition}}

\begin{figure}[t]
    \centering
    \includegraphics[width=1\columnwidth]{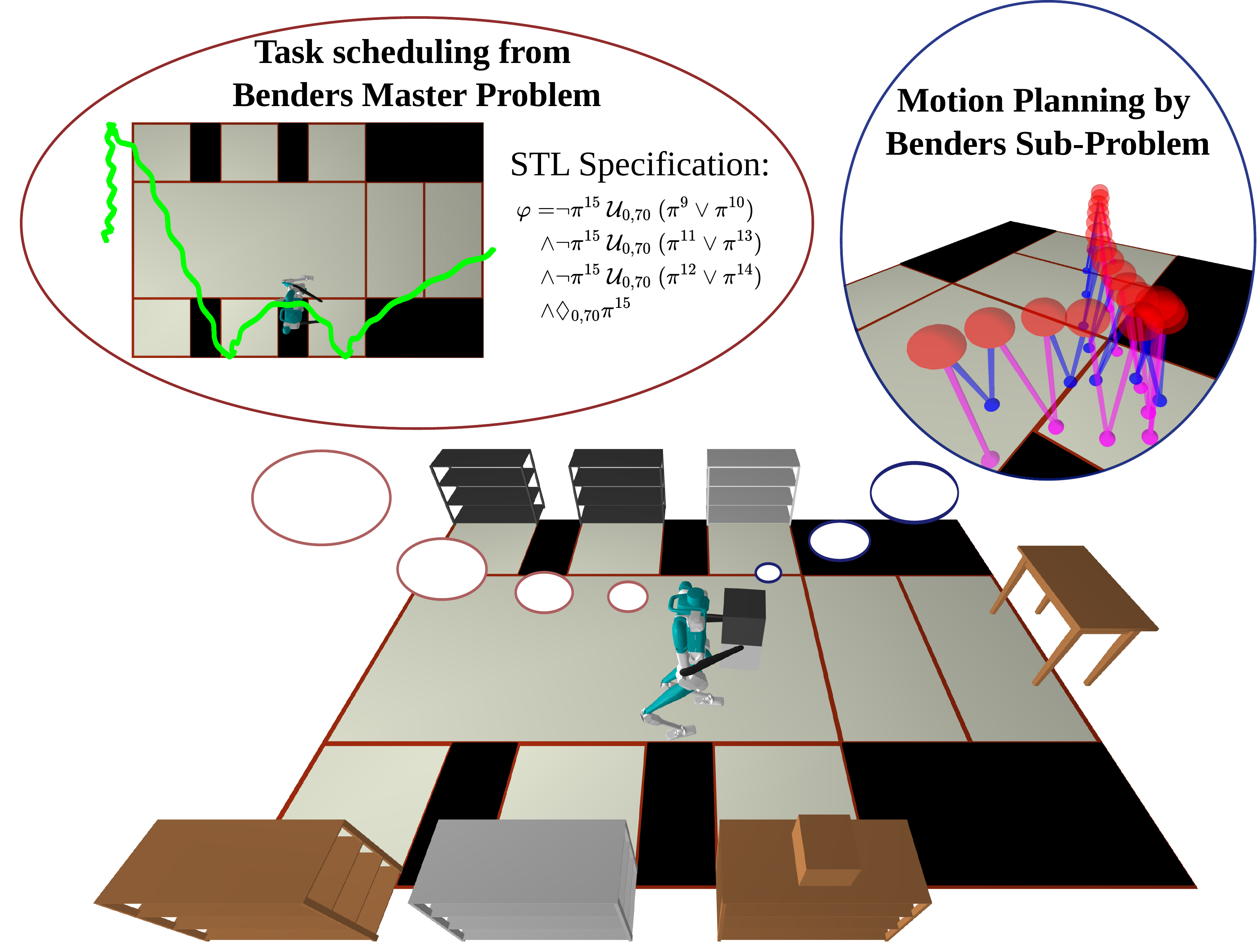}
    \caption{Our bipedal task and motion planning problem is decomposed into a Benders Master Problem, which proposes a task schedule that satisfies the STL task specification, and multiple Benders Sub-Problems, which check the kinematic and dynamic feasibility of the proposed task schedule to be undertaken.}
    \label{fig:pitch}
\end{figure}

\normalsize
\section{Introduction}

As the trend of deploying bipedal robots in environments such as assembly lines and warehouses accelerates, so do expectations for their intelligence to perform a wide range of tasks in a logical and temporally constrained order. A well-suited formal language for specifying such temporal properties, including task ordering and spatial coverage, is Signal Temporal Logic (STL) \cite{maler2004monitoring}. Unfortunately, task planning under STL leveraging optimization-based methods inevitably introduces binary variables into the formulation, making computational overhead intractable when the planning horizon expands \cite{raman2014mpc}. For continuous bipedal motion, this complexity is further exacerbated by the integration of dynamics, kinematics, and discrete task planning under one unified framework.


To this end, many planning formulations tend to decouple trajectory planning from task planning to maintain a manageable computational load \cite{garrett2021integrated, zhou2022reactive} or resort to an oversimplified robot model that sacrifices dynamical feasibility. In the case of bipedal locomotion, this often leads to the use of descriptive reduced-order models, such as single rigid body or centroidal models, which provide sufficient accuracy for whole-body controller tracking. However, these models are unsuitable for direct integration with MIP given their nonlinearity. As a result, simplified models that adequately capture key locomotive features including foot placement and center-of-mass (CoM) velocity, such as the linear inverted pendulum (LIP) \cite{Kajita2003biped} or angular linear inverted pendulum (ALIP) \cite{gong2021angular}, are more commonly adopted in bipedal navigation. In addition, their linear dynamics allows them to be incorporated as constraints in mixed-integer convex program (MICP) formulations, which are widely received for optimization-based task and motion planning (TAMP) under STL \cite{raman2014mpc}. A limitation of this formulation lies in its inability to accommodate nonlinear kinematic reachability constraints capturing foot positions and collision avoidance with complex obstacles.

To balance dynamical feasibility and computational efficiency, this paper exploits the decomposibility of the optimization formulation using Benders Decomposition (BD) \cite{hooker2003logic}. The BD approach decomposes the locomotion TAMP problem into a master problem and multiple subproblems, as illustrated in Fig.~\ref{fig:pitch}, and then solves the TAMP problem by iteratively adding cutting planes to the master problem. The master problem generates candidate task sequences, through solving an MICP with relaxed kinematics and dynamics constraints.
The subproblems, formulated as nonlinear programs (NLPs), verify the feasibility of the proposed task sequences under a full set of constraints. If a subproblem fails to find a feasible solution, feasibility cuts are added to the master problem to eliminate infeasible task sequences in future iterations. On the other hand, if all NLPs are successfully solved, optimality cuts refine the search for improved solutions to the master problem.  This interleaving process of adding feasibility and optimality cuts terminates when a solution meets the optimality criteria.

Another popular decomposition technique -- the Alternating Direction Method of Multipliers (ADMM) \cite{boyd2011distributed} -- has garnered increasing interest in the field of robotics in recent years \cite{zhou2020accelerated, aydinoglu2024consensus, shorinwa2023distributed, shirai2022simultaneous, budhiraja2019dynamics, meduri2023biconmp}. Unlike ADMM, which relies on consensus costs to drive convergence among decoupled subproblems, our proposed BD method introduces a novel cut-shifting mechanism that significantly accelerates convergence. In our MIP formulation, this ensures exact satisfaction of all constraints, and maintains stronger feasibility and optimality guarantees than ADMM, where some constraints are only approximately satisfied.

As one of the first papers introducing BD to accelerate the optimization-based TAMP for bipedal locomotion, this work demonstrates its capability to efficiently compute both an initial feasible solution and the optimal solution, in terms of minimal task completion time, across all benchmark environments. Compared to existing state-of-the-art methods \cite{shirai2022simultaneous, kurtz2022mixed}, our approach enables bipedal robots to plan within reasonable computational time for diverse task specifications, even under scenarios with long planning horizons. Specifically, our contribution can be summarized as follows:
\begin{enumerate}
\item We introduce a novel BD formulation that decomposes the STL-guided TAMP problem for bipedal locomotion into a master problem for task scheduling and subproblems for motion feasibility, enabling efficient planning over long horizons.
\item We propose a novel time-shifted cuts generation mechanism that significantly accelerates the BD solving process and achieve a solving speed at least 20 times faster than other state-of-the-art methods across all benchmark scenarios.
\item We demonstrate the computational advantages of BD for STL-based bipedal TAMP over ADMM, which itself is being designed for STL-guided TAMP problem for the first time. We show that the proposed decomposition methods provide a computationally efficient framework to solve STL-guided TAMP problems with long horizons, such as more than 100 footsteps, and nonlinear kinematics and dynamics.
\end{enumerate}

\begin{figure}[t]
    \centering
    \includegraphics[width=0.8\columnwidth]{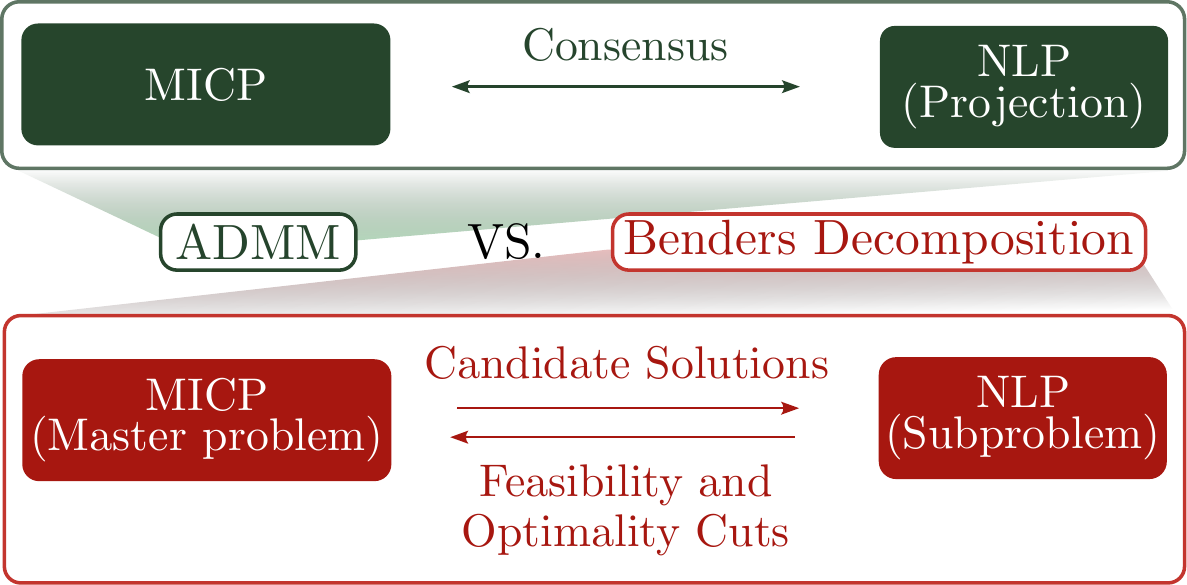}
    \caption{Comparison of the formulation architectures of the Alternating Direction Method of Multipliers (ADMM) and Benders Decomposition.}
    \label{fig:admm_bd_comparison}
\end{figure}


\section{Related Work}
\subsection{Task and Motion Planning under Temporal Logic}
\label{sec:related_TAMP}
Temporal logic-based TAMP ensures formal guarantees for the satisfaction of high-level task specifications, but becomes inherently challenging to solve when integrated with complex system dynamics \cite{gu2025robust, zhao2022reactive, shamsah2024socially}. One way to handle this complexity is to create discrete abstractions of continuous systems. This allows the specification to be enforced through automaton-based synthesis by modeling the system as a finite transition system \cite{ho2022automaton}. However, automaton-based methods often struggle to scale efficiently with increasing model dimensionality and complexity of the specifications \cite{ren2024ltldstar}.

Another paradigm is to employ MICP where system dynamics are typically approximated with double integrator models or piecewise-linear trajectories \cite{sun2022multi}. However, MICP is, in its essence, NP-hard despite its effectiveness in capturing convex dynamics models and temporal logic constraints. Attempts have been made to enhance computational efficiency. 
A smoothing method applies a gradient-based method to solve a relaxed continuous problem, but this comes at the expense of losing completeness guarantees \cite{gilpin2021smooth}. Another approach refines the STL encoding by reducing the number of binary variables to a logarithmic fraction of the original quantity \cite{kurtz2022mixed}. A similar result can be achieved through convex relaxation tightening by leveraging formulations such as graph-of-convex-sets \cite{kurtz2023temporal} or logic network flow \cite{lin2025optimization}.  However, these methods cannot handle the nonlinearities introduced by the robots' embedded kinematic or dynamic constraints. 

\subsection{Locomotion using Mixed-Integer Programming}
\label{sec:related_locomotion}

Mixed-integer programming (MIP) is a powerful tool for addressing contact-rich motion planning problems, particularly in legged locomotion for footstep planning. By introducing integer variables to represent discrete decisions such as foothold selection or gait patterns \cite{aceituno2018simultaneous, zhou2025physically}, MIP effectively consolidates contact events with robot dynamics. Another advantage of MIP in legged locomotion lies in its ability to address long-horizon planning problems. While alternative solutions like NLP often struggle with local minima and convergence issues, MIP leverages the branch-and-bound (B\&B) technique to systematically explore the solution space, ensuring an optimal solution is found if one exists. Specifically, MIP encodes essential physical constraints, including rigid-body dynamics, collision avoidance, and friction cone conditions \cite{deits2014footstep, dai2019global}, through convex sets which are expressed as linear or quadratic constraints. Convex relaxation allows MIP to bypass the computational complexity of mixed-integer nonlinear programs (MINLP), which stems from directly integrating nonlinear constraints with discrete integer variables. However, despite these convex relaxations, solving such formulations remains computationally demanding, especially when planning over long horizons or with many possible contact combinations \cite{marcucci2020warm, ponton2021efficient}.


\subsection{Decomposition Methods for Legged Locomotion Planning}

The complexity of motion planning for legged robots, as discussed in Sec. \ref{sec:related_locomotion}, has motivated numerous strategies aiming for decomposing the problem into more tractable subproblems. A common approach adopts non-iterative hierarchical frameworks \cite{kuindersma2016optimization, shamsah2023integrated, tonneau2018efficient, lin2019optimization}, wherein a high-level planner employs a reduced-order locomotion model to design reference  trajectories, and a low-level planner relies on a full-order model to generate dynamically feasible motions  \cite{norby2022quad}. While this hierarchical paradigm substantially reduces computational workload, it offers no formal guarantee that the high-level plans will be dynamically realizable by the underlying low-level planner. To address this limitation, several approaches have been proposed to introduce feedback or refinement mechanisms, such as through exploiting the differentiability of the low-level planner \cite{le2024fast},  MPC replanning \cite{li2021model}, or  iterative matching between reduced-order and full-order  models \cite{zhou2022momentum}.

A particularly notable iterative technique is the Alternating Direction Method of Multipliers (ADMM) \cite{boyd2011distributed, zhou2020accelerated, aydinoglu2024consensus, shorinwa2023distributed, shirai2022simultaneous, budhiraja2019dynamics, meduri2023biconmp}. ADMM decomposes the full problem into subproblems and iterates among them, using a cost to ensure solution consensus across subproblems. These iterative methods offer computational efficiency through problem decomposition and parallelization. However, they typically lack formal convergence guarantees when applied to locomotion problems involving hybrid dynamics. To address this, extensive weight tuning is often required to achieve approximate solutions that may still violate constraints since consensus is not exactly met.

To address the limitations of existing decomposition approaches such as ADMM for hybrid locomotion, the work of \cite{lin2024accelerate} introduces a Benders Decomposition approach for hybrid model predictive control that provides formal convergence guarantees. Innovative techniques such as cut shifting and warm-starting are introduced to improve computational efficiency. 
This paper extends this method to a new problem domain by accelerating TAMP under temporal logic constraints for long-horizon bipedal locomotion planning.

\section{Preliminaries}

\subsection{Signal Temporal Logic}
\label{[pre:stl}
Consider a discrete time dynamical system of the form 
\begin{equation}
    \boldsymbol x_{k+1} = f(\boldsymbol x_k, \boldsymbol u_k, \boldsymbol z_k)
\end{equation}
where $\boldsymbol x_{k} \in \mathbb{R}^{n_x}$ is the continuous state at step $k$, $\boldsymbol u_{k} \in \mathbb{R}^{n_u}$ is the control input, and $\boldsymbol z_{k} \in \mathbb{R}^{n_z}$ is a vector of binary variables. In this paper, we restrict our focus to bounded-time STL formulas,  where the task specification is defined over a finite time horizon $K$, with $k \in \mathcal K = \{ 1,2, \dots, K\}$. Our STL formulas build on convex predicates, each of which is represented by a system of linear inequalities that describes a convex polytope $\mathcal R_\xi := \{\boldsymbol x_{k}| \boldsymbol{E}_\xi \boldsymbol{x}_k + \boldsymbol{F}_\xi \leq \boldsymbol 0\}$, where $\boldsymbol{E}_\xi \in \mathbb{R}^{n_\xi \times n_x}$,  $\boldsymbol{F}_\xi, \boldsymbol 0 \in \mathbb{R}^{n_\xi}$, and $\xi \in \Xi$ denotes the index of the polytope.


The syntax of STL formulas is recursively defined as
$
\varphi \coloneqq \: \pi \;|\; \neg\varphi \;|\; \varphi_1 \wedge \varphi_2 \;|\; \varphi_1 \vee \varphi_2 \;|\; \Diamond_{[t_1,t_2]} \; \varphi \;|\; \square_{[t_1,t_2]} \; \varphi \;|\; $ $  \varphi_1 \; \mathcal{U}_{[t_1,t_2]}\; \varphi_2
$, 
where $\pi$ is an atomic predicate and $\varphi$, $\varphi_1$, $\varphi_2$ are STL formulas. 
We define that the robot's state $\boldsymbol{x}_k$ satisfies $\pi^\xi_k$ at step $k$, denoted as $\boldsymbol{x}_k \models \pi^\xi_k$, if $\boldsymbol x_k \in \mathcal R_\xi$.
While an atomic predicate only describes a state at a single time step, a formula can characterize a suffix of a trajectory starting from step $k$, expressed as $(\boldsymbol{x}, k)$. The formulas and predicates are connected by boolean operators “and” ($\wedge$), “or” ($\vee$), and “not” ($\neg$), as well as temporal operators ``always" ($\square$), ``eventually" ($\lozenge$), and ``until" ($\mathcal U$) \cite{belta2019formal}. The satisfaction condition of a formula $\varphi$ for a suffix $(\boldsymbol{x}, k)$ is defined inductively in Table \ref{tab:STL_satisfy}.


\begin{table}[t!]
\centering
\caption {\label{tab:STL_satisfy} Validity semantics of Signal Temporal Logic}
\vspace{-0.1in}
\renewcommand{\arraystretch}{1.2}
\begin{tabular}{l c c} \\
\hline
$\boldsymbol{x} \models \varphi$ &$\Leftrightarrow$& $(\boldsymbol{x},0) \models \varphi$  \\
$(\boldsymbol{x},k) \models \pi^\xi$ &$\Leftrightarrow$& $\boldsymbol x[k]\models \pi^\xi_k$ \\
$(\boldsymbol{x},k) \models \neg\varphi$ &$\Leftrightarrow$& $(\boldsymbol{x},k) \not\models \varphi$ \\
$(\boldsymbol{x},k) \models \varphi_1 \wedge \varphi_2$ &$\Leftrightarrow$& $(\boldsymbol{x},k) \models \varphi_1 \wedge (\boldsymbol{x},k) \models \varphi_2$ \\
$(\boldsymbol{x},k) \models \varphi_1 \vee \varphi_2$ &$\Leftrightarrow$& $(\boldsymbol{x},k) \models \varphi_1 \vee (\boldsymbol{x},k) \models \varphi_2$ \\
$(\boldsymbol{x},k) \models \Diamond_{[k_1,k_2]}\varphi$ &$\Leftrightarrow$& $\exists {k^{'}\in[k+k_1,k+k_2]}, (\boldsymbol{x},k^{'}) \models \varphi$ \\
$(\boldsymbol{x},k) \models \square_{[k_1,k_2]}\varphi$ &$\Leftrightarrow$& $\forall {k^{'}\in[k+k_1,k+k_2]}, (\boldsymbol{x},k^{'}) \models \varphi$\\
$(\boldsymbol{x},k) \models {\varphi_1}\mathcal{U}_{[k_1,k_2]}{\varphi_2}$ &$\Leftrightarrow$& $\exists {k^{'}\in[k+k_1,k+k_2]}, (\boldsymbol{x},k^{'}) \models \varphi_2$ \\
&& $\wedge \ \forall {k^{''}\in[k+k_1,k^{'}]} (\boldsymbol{x},k^{''}) \models \varphi_1$ \\ 
\hline
\end{tabular}
\end{table}

\subsection {STL Specifications as MICP Constraints}
\label{sec:stlasmicp}
When an STL specification is recursively composed from predicates and formulas using Boolean and temporal operators, it naturally induces a tree structure \cite{wolff2014optimization, raman2014mpc}, where the root node represents the complete specification and the leaf nodes correspond to the individual predicates. Each node except the root node is an immediate suboperand of its parent node, which combines its children through an operator. In particular, temporal operators can be reformulated as Boolean operators applied over sequences of operands, indexed by discrete time steps within a specified temporal window, as indicated in Table I. For instance, $\square_{[k_1, k_2]}\varphi$ and $\lozenge_{[k_1, k_2]}\varphi$ can be respectively translated to $\wedge_{k \in [k_1, k_2]}(\boldsymbol{x}, k) \models \varphi$ and $\vee_{k \in [k_1, k_2]}(\boldsymbol{x}, k) \models \varphi$. This reformulation results in connectives between parent and child nodes consisting solely of conjunctions and disjunctions. 

To encode an STL specification as the constraints of an MICP formulation, we assign each predicate on the leaf node with a binary variable $z^{\pi_k^\xi} \in \boldsymbol z_k$ such that $z^{\pi_k^\xi} =1  \Rightarrow \boldsymbol{x}_k \models \pi^\xi_k $. This implication relationship can be encoded through the big-M method \cite{raman2014mpc}:
\begin{equation}
\boldsymbol{E}_\xi \boldsymbol{x}_k + \boldsymbol{F}_\xi \leq M(\boldsymbol 1 - z^{\pi_k^\xi} \boldsymbol 1)  
\label{eqn:bigm}
\end{equation}
where $\boldsymbol 1 \in \mathbb R^{n_\xi}$ and $M\gg 1$ is a scalar. 

Moreover, a binary variable is designated to each formula $\varphi$ on the non-leaf nodes and is denoted as $z^\varphi \in \mathbb {B}$. Here we connect the binary variables of the operands with parent-child relations through the following inequality constraints:
\begin{itemize}
    \item For a formula that aggregates its immediate suboperands through conjunction connective: $\varphi = \wedge_{i=1}^m \varphi_i$, assuming $\varphi_i$ is either a formula or a predicate of the child nodes:
    \begin{subequations}
    \begin{align}
    z^\varphi &\leq z^{\varphi_i}, \ \ i = 1, \ldots , m \label{eqn:tree1a} \\
    z^\varphi &\geq 1-m+\sum_{i=1}^m z^{\varphi_i} \label{eqn:tree1b}
    \end{align}
    \label{eqn:tree1}
    \end{subequations}
    \vspace{-3mm}
    \item Similarly, for a formula with disjunctive connective:  $\varphi = \vee_{i=1}^n \varphi_i$:
    \vspace{-3mm}
    \begin{subequations}
    \begin{align}
    z^\varphi &\geq z^{\varphi_i}, \ \ i = 1, \ldots , n \label{eqn:tree2a} \\
    z^\varphi &\leq \sum_{i=1}^n z^{\varphi_i} \label{eqn:tree2b}
    \end{align}
    \label{eqn:tree2}
    \end{subequations}
    \vspace{-4mm}
\end{itemize}
These constraints together with \eqref{eqn:bigm} inherently enforce the satisfaction of the full task specification $\varphi_0$ provided that $z^{\varphi_0}=1$.
We define a constraint
\begin{equation}
    \mathcal{C}_{\vee, \wedge}(\mathbf{Z}) \leq 0
    \label{eqn:connectives}
\end{equation}
where $\mathbf Z \in \mathbb B^{|\mathbf{Z}|}$ encompasses $\boldsymbol z_k$ for all steps, as well as $z^\varphi$ for every formula on the non-leaf nodes. We let \eqref{eqn:connectives} incorporate all the inequality constraints on the non-leaf nodes in \eqref{eqn:tree1} and \eqref{eqn:tree2}, along with $z^{\varphi_0} \leq 1$ and $z^{\varphi_0} \geq 1$. These linear inequalities in \eqref{eqn:connectives} also lend themselves to the MICP formulation.

\subsection{Reduced-order Bipedal Dynamics Model}

\begin{figure}[t]
\centering
\includegraphics[width=1\columnwidth]{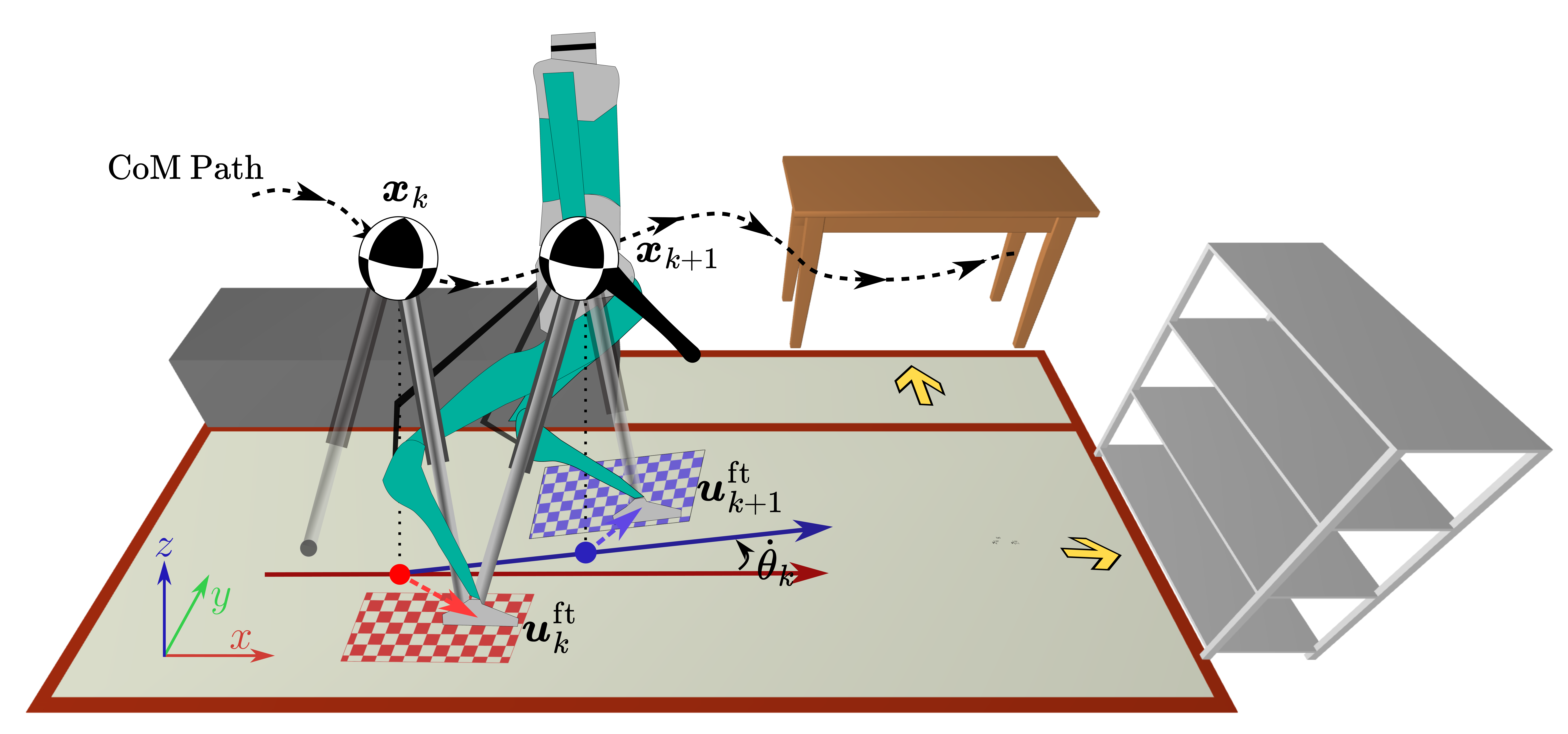}
    \caption{An illustration of the simplified dynamics of the bipedal robot Digit using a LIP model. The colored checkered patterns approximate the reachable set of foothold locations in the $xy$-plane. Red boxes indicate the boundaries of each convex region, and yellow arrows highlight points of interest along with their associated heading angles.}
    \label{fig:footstep_constraints}
\end{figure}

Bipedal locomotion process can be approximated as a 3-D Linear Inverted Pendulum (LIP) model that represents a point mass atop a massless prismatic leg \cite{Kajita2003biped}, as illustrated in Fig. \ref{fig:footstep_constraints}. This work adopts a discrete LIP formulation where the state variables are defined at the contact moment of each footstep\footnote{All steps referred in the following sections denote walking footsteps.}. At each contact, the LIP model consists of a 5-dimensional continuous state vector $\boldsymbol{x}_k = [\boldsymbol p_k; \boldsymbol  v_k; \theta_k] \in \mathbb{R}^5$ comprising a 2-D center-of-mass (CoM) position vector $\boldsymbol p_k = [x_k, y_k]^\top$, a 2-D CoM velocity vector $\boldsymbol  v_k = [\dot x_k, \dot y_k]^\top$, and an orientation angle $\theta_k$ of the point mass with respect to an inertia frame. The point mass on top of the leg is assumed to move in a horizontal plane located at a constant height $H$, and therefore the vertical CoM position is omitted from the state. The control input is defined as $\boldsymbol{u}_k = [ \boldsymbol u^{\rm ft}_k; \dot \theta_k] \in \mathbb{R}^3$, where $\boldsymbol u^{\rm ft}_k = [ u_{k}^x,u_{k}^y]^\top$ is the 2-D foodhold location of the stance foot relative to the CoM expressed in the inertia frame, and $\dot\theta_k$ is the angular velocity.

The continuous-time evolution of the LIP model along the $x$ axis is governed by $\ddot x = -(g/H)u^x$, where $g$ is the gravitational acceleration constant. Identical dynamics applies to the motion along the $y$ axis. These expressions can be integrated into discrete representation in 2-D Euclidean space \cite{narkhede2022sequential}:
\begin{subequations}
\begin{align}
\begin{bmatrix} x_{k+1} \\ \dot{x}_{k+1} \end{bmatrix} &= \begin{bmatrix} 1 & \frac{1}{\omega} \sinh(\omega T) \\ 0 & \cosh(\omega T)\end{bmatrix} \begin{bmatrix}  x_{k} \\ \dot{x}_k \end{bmatrix} + \begin{bmatrix} 1 - \cosh(\omega T)\\ -\omega \sinh(\omega T)  \end{bmatrix} u_{ k}^x\\
\begin{bmatrix} y_{k+1} \\ \dot{y}_{k+1} \end{bmatrix} &= \begin{bmatrix} 1 & \frac{1}{\omega} \sinh(\omega T) \\ 0 & \cosh(\omega T)\end{bmatrix} \begin{bmatrix}  y_{k} \\ \dot{y}_k \end{bmatrix} + \begin{bmatrix} 1 - \cosh(\omega T)\\ -\omega \sinh(\omega T)  \end{bmatrix} u_{k}^y \\
\theta_{k+1} &= \theta_k + T\dot\theta_k
\end{align}
\label{eqn:LIP}
\end{subequations}
where $\omega = \sqrt{g/H}$ and $T$ is the time interval of two consecutive
walking steps. $\boldsymbol{x}_{k+1} = [\boldsymbol p_{k+1}; \boldsymbol  v_{k+1}; \theta_{k+1}]$ describes the robot's state at the following contact moment.

To ensure kinematic feasibility, the following constraints are imposed to delineate feasible regions for footstep placement, represented as a rectangular bounding box within the foot reachable set and depicted as checkered patterns in Fig. \ref{fig:footstep_constraints}:
\begin{gather}
\boldsymbol{lb} \leq \boldsymbol{R}(\theta_k)^\top \boldsymbol u^{\rm ft}_k \leq \boldsymbol{ub}
\label{eqn:kinematics}
\end{gather}
where $\boldsymbol{R}(\theta_k) \in SO(2)$ denotes the coordinate transformation from the global to the local inertia frame of the point mass. The vectors $\boldsymbol{lb}$ and $\boldsymbol{ub}$ specify the lower and upper bounds that define the feasible foothold region. These bounds are empirically set to the values in Table \ref{Tab:parameters}, based on the nominal stride length at normal walking gait. With a slight generalization of this formulation, \eqref{eqn:kinematics} can also be employed to find collision-free footholds.

To ensure stability, we also prohibit the robot from exhibiting high linear and angular velocities simultaneously:
\begin{equation}
\parallel\boldsymbol v_k\parallel/ v_{\rm max} + |\dot\theta_k|/(\omega T) \leq 1
\label{eqn:stability}
\end{equation}
where $\mid\cdot\mid$ and $\parallel \cdot \parallel$ represent the L1 and L2 norms, and $v_{\rm max}$ is the maximum velocity in any direction. This constraint is tighter than independent limits on linear and angular velocities, respectively.
We also design the locomotion-related objective functions to penalize large shifts in orientation, guide each foot toward its desired location on the lateral axis, and discourage excessive lateral velocity:
\begin{align}
\begin{aligned}\label{eqn:loco-cost}
f^{\rm loco}_k = \omega_1 {\dot\theta_k}^2 &+ \omega_2(\boldsymbol e_y\boldsymbol{R}(\theta_k)^\top\boldsymbol u^{\rm ft}_k-l^{\rm des}_k)^2 \\ 
&+ \omega_3(\boldsymbol e_y\boldsymbol{R}(\theta_k)^\top \boldsymbol v_k)^2
\end{aligned}
\end{align}
where $l^{\rm des}_k$ is the desired lateral distance of feet from the sagittal plane, and $\boldsymbol e_y = [0,1]$ is the selection vector for the lateral component. $\omega_1$, $\omega_2$, and $\omega_3$ are the relative weights of each term. 

The inclusion of the rotation matrix introduces nonlinearity into the formulation, which becomes an MINLP problem. Rather than addressing the full complexity of an MINLP, the next section proposes a more computationally tractable alternative through Benders Decomposition. 

\subsection{Benders Decomposition}
\label{sec:BD}

Benders Decomposition is an effective tool to address \textit{complicating variables} in an optimization problem. Complicating variables are defined as variables that, when fixed, simplify the remaining problem substantially \cite{geoffrion1972generalized}. As shown in Fig. \ref{fig:admm_bd_comparison}, the BD method decomposes the original problem into a \textit{Benders Master Problem} (BMP) that solves the complicating variables and a \textit{Benders Sub-Problem} (BSP) that solves the remaining variables. The algorithm proceeds iteratively: the BMP proposes candidate solutions for the complicating variables, which are then anchored in the BSP to determine optimal values for the remaining variables. If the BSP finds a feasible solution, optimality cuts are added to the BMP to guide it towards finding better candidate values for the complicating variables. If the BSP proves infeasible, feasibility cuts are added to the BMP to remove the infeasible candidate solutions from future iterations. This process continues until reaching optimality criteria or proving infeasibility for the original problem.

Mathematically, consider the optimization problem:
\begin{align}
    \begin{split}
    \minimize_{\boldsymbol{\gamma} \in \mathcal{D}_1, \boldsymbol{\eta} \in \mathcal{D}_2} \quad & F(\boldsymbol{\gamma}) + G(\boldsymbol{\eta}) \\
    \text{s.t.} \quad & H(\boldsymbol{\gamma}, \boldsymbol{\eta}) \leq \boldsymbol{0}
    \end{split}
    \label{eqn:BDformulation}
\end{align}
where $\boldsymbol{\eta}$ represents the complicating variables. By fixing $\boldsymbol{\eta}$, denoted $\boldsymbol{ \widehat{\eta}}$, we obtain a BSP formulation that is typically more computationally tractable:
\begin{subequations}
    \begin{align}
    \minimize_{\boldsymbol{\gamma} \in \mathcal{D}_1} \quad & F(\boldsymbol{\gamma}) \label{eqn:subproblem_fobj} \\
    \text{s.t.} \quad & H(\boldsymbol{\gamma} \,|\, \boldsymbol{ \widehat{\eta}}) \leq \boldsymbol{0}
\end{align}
\label{eqn:subproblem_general}
\end{subequations}
Note that $G(\boldsymbol{ \widehat{\eta}})$ becomes a constant and is removed from the objective function \eqref{eqn:subproblem_fobj}. 
Let $v(\boldsymbol{ \widehat{\eta}})$ denote the optimal cost of \eqref{eqn:subproblem_general}, parameterized by $\boldsymbol{ \widehat{\eta}}$. If   \eqref{eqn:subproblem_general} is proven infeasible, we rule out $\boldsymbol{ \widehat{\eta}}$ from the candidate set of $\boldsymbol{\eta}$. Define a set of $\boldsymbol{\eta}$ that makes the subproblem feasible for $\boldsymbol{\gamma}$ as:
\begin{equation}
    \mathcal{V} = \{\boldsymbol{\eta} \in \mathcal{D}_2 | \;\exists \: \boldsymbol{\gamma} \in \mathcal{D}_1 \text{ s.t. } H(\boldsymbol{\gamma} \,|\, \boldsymbol{\eta}) \leq \boldsymbol{0}\}
\end{equation}
Through Benders Decomposition, the optimization problem in \eqref{eqn:BDformulation} can be solved by first solving the BMP:
\begin{align}
\begin{split}
    \minimize_{\boldsymbol{\eta} \in \mathcal{D}_2} \quad & v(\boldsymbol{\eta}) + G(\boldsymbol{\eta}) \\
    \text{s.t.} \quad & \boldsymbol{\eta} \in \mathcal{V}
\end{split}
\label{eqn:masterproblem_general}
\end{align}
and then solving the subproblem \eqref{eqn:subproblem_general} with the obtained $\boldsymbol{\widehat{\eta}}$. Unfortunately, writing down $v(\boldsymbol{\eta})$ and $\mathcal{V}$ completely in a closed form is difficult in general. Therefore, the iterative process of adding cutting planes can be viewed as constructing the closed-form representations on demand. The complete procedure of BD is outlined in Algorithm~\ref{algorithm_complete_benders}.

\begin{algorithm}[t!]
\caption{Benders Decomposition}
\label{algorithm_complete_benders}
\textbf{Input:} Problem formulation\\
\While{not converged}{
    \textbf{Step 1:} Solve BMP \eqref{eqn:masterproblem_general}\\
    \quad Obtain candidate solution $\boldsymbol{\eta}^{(i)}$\\
    
    \textbf{Step 2:} Solve BSP \eqref{eqn:subproblem_general} with $\boldsymbol{\eta} = \boldsymbol{\eta}^{(i)}$\\
    
    \If{Subproblem is feasible}{
        Obtain optimal solution $\boldsymbol{\gamma}^{(i)}$\\
        Add optimality cut to BMP
    }
    \Else{
        Add feasibility cut to BMP to exclude $\boldsymbol{\eta}^{(i)}$
    }
}
\textbf{Return:} Optimal solution $(\boldsymbol{\gamma}^*, \boldsymbol{\eta}^*)$
\end{algorithm}
\section{STL-Based TAMP of Bipedal Locomotion using Benders Decomposition}

\subsection{Problem Formulation}
\label{sec:problem_formulation}
We begin this section by formulating the robot’s working environment and the STL predicates through a motivating example. Consider a bipedal robot operating on a factory floor as shown in Fig. \ref{fig:footstep_constraints}, where the robot’s configuration space $\mathcal{W}$ is partitioned into a set of convex polytopes $\mathcal{R}_\mu \subseteq \mathcal{W}$, indexed by $\mu \in \Xi_{\rm reg}$, where $\Xi_{\rm reg}$ includes the indices for all regions. Constraint \eqref{eqn:bigm} is employed to ensure that both the CoM and the foothold reside within a convex region whenever the corresponding predicate is satisfied. By explicitly requiring that
\begin{equation} 
    \sum_{\mu \in \Xi_r} z^{\pi^\mu_k} = 1
    \label{eqn: oneregion}
\end{equation}
we confine the robot to stay within one region at each step $k$. Convex region boundary constraint through \eqref{eqn:bigm} and mutually exclusive selection constraint \eqref{eqn: oneregion}
collectively guarantee a collision-free trajectory.
    

We also identify a set of standalone locations on the map, denoted as $\mathcal R_\nu \in \mathcal W$, $\nu \in \Xi_{\rm poi}$, referred to as ``points of interest" (PoI). These distinct spots represent workbenches or pallet totes where the robot must reach to perform specific tasks. Each PoI also specifies the desired orientation of the robot, and the robot is required to remain completely stationary to allow the upper body to perform tasks like picking up boxes or totes.
These additional requirements concerning orientation and static pose are also encoded as linear inequalities and incorporated in $\mathcal R_\nu$. We assume that the tasks assigned to the robot involve accessing PoIs while either passing through or avoiding certain regions under some logical and temporal conditions.

Based on the aforementioned formulation of the environment, we define two vectorized binary variables: 
\begin{subequations}
\begin{align}
\boldsymbol{z}^{\pi^\mu} &= \{z^{\pi_k^\mu} \mid k \in \mathcal K, \mu \in \Xi_{\rm reg}\} \\
\boldsymbol{z}^{\pi^\nu} &= \{z^{\pi_k^\nu} \mid k \in \mathcal K, \nu \in \Xi_{\rm poi}\} \label{def:task_schedule}
\end{align}
\end{subequations}
where $\boldsymbol{z}^{\pi^\mu}$ specifies the robot's 2D CoM path through the convex regions, and
$\boldsymbol{z}^{\pi^\nu}$ encodes the task schedule which indicates the timings and the positions of visited PoIs.
The entire index set is a union of indices of both the regions and the points $\Xi = \Xi_{\rm reg} \cup \Xi_{\rm poi}$. 

In addition to generating a collision-free locomotion trajectory, we also aim to minimize the time span required to complete the entire task. This can be realized through minimizing the following term: 
\begin{equation}
f^{\rm time} = \sum_{k=0}^K e^k\left(\sum_{\nu \in \Xi_{\rm poi}} z^{\pi_k^\nu} \right)
\end{equation}
where we intentionally apply a heavier cost via the exponential cost weight, if the robot reaching a PoI later in the course.
The complete optimization formulation becomes:
\begin{align*}
\underset{\mathbf{X}, \mathbf{U}, \mathbf{Z}}{\text{minimize}}& \quad f^{\rm time} + \sum_{k=1}^{K} f^{\rm loco}_k \nonumber \\
\text{s.t.} 
& \quad
\begin{cases}
\left.\begin{aligned}
&\eqref{eqn:bigm}, \ \forall k\in \mathcal K \;\; \forall \xi \in \Xi; \\ 
&\eqref{eqn:connectives}; \\ 
&\eqref{eqn: oneregion}, \ \forall k \in \mathcal K;
\end{aligned}\right\}\ \ \ \text{(STL-related Cons.)} \\
\eqref{eqn:LIP}\ \eqref{eqn:kinematics}\ \eqref{eqn:stability},\ \forall k \in \mathcal K \backslash K ; \ \ \  \text{(Locomotion Cons.)}
\end{cases}
\vspace{-3mm}
\end{align*}
where $\mathbf {X} = \{\boldsymbol{x}_k |k\in\mathcal K\}$, $\mathbf {U} = \{\boldsymbol{u}_k |k\in\mathcal K\backslash K\}$, and $\mathbf Z$ is defined in Sec. \ref{sec:stlasmicp}. The locomotion-related cost function $f^{\rm loco}_k$ is introduced in \eqref{eqn:loco-cost}.
Due to the nonconvexity of locomotion constraints in the problem formulation, we define the optimal solution as the set of decision variables that yields the minimal value of $f^{\rm time}$.

\subsection{Decomposition Strategy}
Since the full formulation combining discrete task planning with continuous motion planning under nonlinear constraints is computationally prohibitive, we propose using Benders Decomposition (BD) to solve this problem more efficiently. While BD is a well-established method in the optimization community for solving large-scale MIPs, it has seen limited applications in robotics field. In this paper, we introduce BD to bipedal TAMP and extend it with novel time-shifted cutting plane generation techniques that significantly accelerate convergence. Based on our problem definition in Sec. \ref{sec:problem_formulation}, we decompose the overall formulation into a Benders Master Problem (BMP) and multiple Benders Sub-Problems (BSPs).

\textit{1) Benders Master Problem:} We formulate the BMP to balance model fidelity and computational speed, such that the relaxed model remains sufficiently accurate to provide instructive priors toward the BSPs, while maintaining computational efficiency over solving the full formulation. The constraints of BMP include the LIP model dynamics 
and the STL-related constraints, while relaxing the foot reachability constraint   \eqref{eqn:kinematics} and stability constraint   \eqref{eqn:stability}. 

We simplify \eqref{eqn:kinematics} to a fixed rectangular region independent of body orientation and the left-right gait pattern:
\begin{equation}
|u_{k}^{x}| \leq r_{\rm max}, \quad |u_{k}^{y}| \leq r_{\rm max}
\label{eqn:circular_reach}
\end{equation}
where $r_{\rm max}$ is the maximum reachable distance. These bounds are chosen to enclose the union of all orientation-dependent reachable regions, ensuring that no feasible solutions are excluded from the relaxation. This approximation also removes the nonlinear rotation matrix $\boldsymbol{R}(\theta_k)$ in \eqref{eqn:kinematics}, thereby avoiding the formulation of MINLP or induction of additional binary variables for piecewise linear approximation of trigonometric functions, as in \cite{deits2014footstep}. Similarly, we relax \eqref{eqn:stability} by imposing independent bounds on the linear and angular velocities:
\begin{equation}
|\dot x_k| \leq v_{\rm max}, \quad |\dot y_k| \leq v_{\rm max}, \quad |\dot \theta_k| \leq \omega T
\label{eqn:velocity_bound}
\end{equation}
to avoid the nonlinear L1 and L2 norms. Our MICP formulation for the BMP is therefore simplified to:
\begin{align*}
\underset{\mathbf{X}, \mathbf{U}, \mathbf{Z}}{\text{minimize}} \quad  f^{\rm time} \nonumber
\begin{aligned}
\quad \quad \quad \text{s.t.} 
\quad
\begin{cases}
\eqref{eqn:bigm}, \ \forall k\in \mathcal K \;\; \forall \xi \in \Xi; \\ 
\eqref{eqn:connectives}; \\ 
\eqref{eqn: oneregion}, \ \forall k \in \mathcal K; \\
\eqref{eqn:LIP}\ \eqref{eqn:circular_reach}\ \eqref{eqn:velocity_bound}, \ \forall k \in \mathcal K \backslash K; 
\end{cases}
\end{aligned}
\end{align*}
The BMP outputs a feasible task schedule $z^{\pi_k^{\nu}}$, from which we can extract a sequence of PoIs $ \mathcal{R}_{\nu_1}, \mathcal{R}_{\nu_2}, \ldots, \mathcal{R}_{\nu_n} $ to be visited at specific steps $ k_1, k_2, \ldots, k_n $, by identifying the entries in the schedule where $z^{\pi_{k_i}^{\nu_i}} = 1$. While this task schedule satisfies the STL task specification, it does not guarantee locomotion kinematic feasibility or stability. 

\textit{2) Benders Sub-Problem:} The BSP decomposes the overall motion planning problem into a series of independent trajectory optimization problems between consecutive PoIs. A fixed task schedule $\boldsymbol{z}^{\pi^\nu}$ provided by the BMP includes a definite, chronologically ordered sequence and timing for arriving at the PoIs $ \mathcal{R}_{\nu_i}$ for $i \in \mathcal I= \{ 1, 2, \ldots, n\}$. We also let $k_0 = 0$ and identify a polytope $\mathcal R_{\nu_0}$ such that the robot's initial state $\boldsymbol x_0 \in \mathcal R_{\nu_ 0}$. In our case, each individual problem of the BSPs generates a trajectory from $\boldsymbol{x}_{k_{i-1}} \in \mathcal{R}_{\nu_{i-1}}$ to $\boldsymbol{x}_{k_i} \in \mathcal{R}_{\nu_i}$, for $i \in \mathcal I$. The BSPs solve these problems using LIP model dynamics \eqref{eqn:LIP} with exact foot kinematic reachability \eqref{eqn:kinematics} and stability constraints \eqref{eqn:stability}. Since $\boldsymbol{z}^{\pi^\nu}$ already satisfies the STL specification, constraint \eqref{eqn:connectives} is omitted in the BSP formulations.


In this work, we model the BSP as an NLP for computational efficiency. Instead of restricting the robot to a set of convex regions, we enforce collision avoidance using smooth constraints based on signed distance functions. For obstacles of arbitrary shape, we split them into polygons $\mathcal{R}_{\lambda}$, $\lambda \in \Xi_{\rm obs}$, from which we compute the signed distance from both the CoM position and the foothold position, to each edge of the obstacle. The collision avoidance constraint is expressed as a log-sum-exp softmin approximation \cite{rabiee2025softminimum}:
\begin{align}
H(\boldsymbol{p}) = h_{\min} - \frac{1}{\alpha} \log\left(\sum_{j} e^{(-\alpha (h_j(\boldsymbol{p}) - h_{\min}))}\right) + \epsilon \leq 0 \label{eqn:softmin_collision}
\end{align}
    where $h_j(\boldsymbol{p})$ represents the signed distance of either CoM or foothold position $\boldsymbol p$, to the $j$-th edge of the obstacle. $h_{\min}$ is the minimum signed distance among all edges, $\alpha$ controls the approximation sharpness, and $\epsilon > 0$ provides a safety margin. This smooth constraint replaces \eqref{eqn:bigm} and does not require binary variables for explicit region assignment, while still guaranteeing a collision-free trajectory. With this NLP formulation, each trajectory segment can be solved in under one second. The full formulation for the BSP is:
\begin{align*}
&(\text{For each motion segment from } \mathcal{R}_{\nu_{i-1}} \text{ to } \mathcal{R}_{\nu_{i}}, \ i\in \mathcal I )\\
&\underset{\mathbf{X}, \mathbf{U}}{\text{minimize}} \quad \sum_{k=k_{i-1}}^{k_{i}} f^{\rm loco}_k
\quad\quad
\begin{cases}
\eqref{eqn:softmin_collision}, \ \forall \lambda \in \Xi_{\rm obs}\; \forall k \in [k_{i-1}, k_{i}]; \\
\eqref{eqn:LIP}\ \eqref{eqn:kinematics}\ \eqref{eqn:stability}, \ \forall k\in [k_{i-1}, k_{i}-1];\\
\boldsymbol{x}_{k_{i-1}} \in \mathcal{R}_{\nu_{i-1}}, \ \boldsymbol{x}_{k_{i}} \in \mathcal{R}_{\nu_{i}};
\end{cases}
\end{align*}
\subsection{Cutting Plane Generation}
\label{sec:cutting plane}
As discussed in Sec. \ref{sec:BD}, cutting planes are generated and added to the BMP after each iteration to prune infeasible or suboptimal solutions.
When the BSPs detect constraint violation under a fixed task schedule $\widehat{\boldsymbol{z}}^{\pi^\nu}$, a feasibility cut is added to the BMP to remove the proposed task schedule from future iterations. If at least one feasible task schedule exists in the BMP and assuming feasibility cuts only remove provably infeasible task schedules, the BD algorithm will eventually find a solution. This condition holds when the BSPs are formulated as convex programs or MIPs, where infeasibility certificates provide rigorous guarantees. However, our BSPs are formulated as NLPs, and therefore, this condition does not hold theoretically due to the problem's non-convexity. Even so, modern NLP solvers with proper initialization can reliably detect true infeasibility in practice, particularly for our structured motion planning problems with smooth collision avoidance constraints. We presume that the condition above largely holds for our setting.

Given an infeasible solution $\widehat{\boldsymbol{z}}^{\pi^\nu}$, a logic-based Benders cut \cite{codato2006combinatorial} is applied to eliminate it from the BMP:
\begin{equation}
    \sum_{\forall k,\nu \, \text{s.t.} \, \widehat{z}^{\pi_k^\nu} = 1} (1-z^{\pi_k^\nu}) + \sum_{\forall k,\nu \, \text{s.t.} \, \widehat{z}^{\pi_k^\nu} = 0} z^{\pi_k^\nu} \geq 1
\label{eqn:feas_cut}
\end{equation}
%
Specifically, the cut includes the term $1-z^{\pi_k^\nu}$ if the infeasible solution has $\widehat{z}^{\pi_k^\nu} = 1$, and includes $z^{\pi_k^\nu}$ if $\widehat{z}^{\pi_k^\nu} = 0$. 
This cut enforces that at least one binary variable in any future solution must differ from the infeasible task schedule $\widehat{\boldsymbol{z}}^{\pi^\nu}$, thereby avoiding the generation of the same infeasible solution by the BMP. Consider a simple example as follows:

\begin{example}
An environment contains 3 PoIs, where $\mathcal I = {1,2,3}$, and the assigned task spans over a 2-step horizon (i.e., $K = 2$). Assume an infeasible task schedule $\widehat{\boldsymbol{z}}^{\pi^\nu} = [1, 0, 0; 0, 0, 1]$, where the first row represents reaching PoI 1 at step $k=1$ and the second row represents reaching PoI 3 at step $k=2$.
Following~\eqref{eqn:feas_cut}, the feasibility cut induced by $\widehat{\boldsymbol{z}}^{\pi^\nu}$ is given as:
\begin{equation*}
(1-z^{\pi^{\nu_1}_1}) + (1-z^{\pi^{\nu_3}_2}) + z^{\pi^{\nu_2}_1} + z^{\pi^{\nu_3}_1} + z^{\pi^{\nu_1}_2} + z^{\pi^{\nu_2}_2} \geq 1  \end{equation*}
\end{example}





On the other hand, an optimality cut is added to the BMP if all BSPs find feasible solutions. For BSPs formulated as convex programs, the optimality cut is typically constructed by leveraging duality theory to provide rigorous lower bounds on the objective function \cite{geoffrion1972generalized}. Because of the non-convexity of BSPs in our formulation, we generate optimality cuts in a manner similar to how we generate feasibility cuts. Specifically, we keep track of the best solution that yields the earliest finish time. If the BSP finds a solution that finishes later than the current best, we treat this solution as infeasible and add a cut identical to \eqref{eqn:feas_cut}. This enables the BMP to only explore different schedules that provide lower cost for $f^{\rm time}$.

A key principle to improve BD efficiency is to ``learn as much as possible from one failure." Instead of adding a single cut at a time, we aim to generate multiple strong cuts after each iteration.
Because we intend for the task schedule that finishes all tasks earlier in time, we could identify the strong cuts in each infeasible BMP trial to rule out other schedules that attempts to reach target PoIs earlier than the infeasible solution discovered. Such solutions appear attractive to the relaxed BMP model but are guaranteed to be infeasible for the BSP. The intuition is straightforward: if a schedule that starts from an initial position and visits a target PoI $\nu$ at step $k$ (i.e., $\widehat{z}^{\pi^\nu_k} = 1$) is infeasible due to dynamics or kinematics constraints, then reaching the same target at an earlier step $k'< k$ (i.e., $\widehat{z}^{\pi^\nu_{k'}} = 1$) must also be infeasible, as it imposes an even stricter time constraint for earlier arrival. In this paper, we design these cuts, termed \textit{time-shifted cuts}, to preemptively eliminate these infeasible sets of binary variables $\widehat{\boldsymbol{z}}^{\pi^\nu}$, which can significantly reduce the total number of BD iterations required for convergence. First introduced in \cite{lin2024accelerate}, a preliminary version of time-shifted cuts is applied to hybrid model predictive control with quadratic subproblems. In this paper, we extend this approach to handle nonlinear subproblems in bipedal TAMP. 

The general idea behind the time-shifted cuts is to move the infeasible task schedule to an earlier time window by sliding each task assignment forward (e.g., let the shifting steps be $s \in \{1, 2, \ldots, s_{\max}\}$, then we have $\widehat{z}^{\pi^\nu_k} \mapsto \widehat{z}^{\pi^\nu_{k-s}}$ for $k = s, s+1, \ldots$), and padding the remaining binary variables with zeros. This ensures that if a PoI $\pi^\nu_k$ is unable to be reached at step $k$, then it also cannot be reached at any earlier step  $1, \ldots, k-1$. 
However, certain conditions must be met to guarantee that time-shifted cuts only remove strictly infeasible task schedules, which are stated and formally proved in Appendix~\ref{proof_cut_shifting}. 
The time-shifted cuts take the form:
\begin{align}
\forall s \in &\{1, 2, \ldots, s_{\max}\}: \nonumber \\
&\sum_{\substack{\forall k,\nu \:\text{s.t.}\: \widehat{z}^{\pi_{k+s}^\nu} = 1 \\ \text{where}\: k+s \leq K}} (1-z^{\pi_{k}^\nu}) + \sum_{\substack{\forall k,\nu \:\text{s.t.}\: \widehat{z}^{\pi_{k+s}^\nu} = 0 \\ \text{where}\: k+s \leq K}} z^{\pi_{k}^\nu} \geq 1
\label{eqn:shifted_cut2}
\end{align}
%
where $s_{\max}$ is the maximum allowable shifting steps determined by the condition that if constraint~\eqref{eqn:bigm} is infeasible under the initial condition $\boldsymbol{x}_0$, it should remain infeasible after shifting, i.e., under $\widehat{\boldsymbol{z}}^{\pi^\nu_s}$ for all $s \leq s_{\max}$. 

\begin{example}[continued]
Now we extend the horizon in \textit{Example 1} to $K=3$ and assume that the infeasible schedule is  $\widehat{\boldsymbol{z}}^{\pi^\nu} = [1, 0, 0; 0, 0, 0; 0, 0, 1]$. Here the robot starts from PoI 1 at $k = 1$ but reaches PoI 3 at $k = 3$. Apart from the original cut~\eqref{eqn:feas_cut}, we add time-shifted cut~\eqref{eqn:shifted_cut2} expressed as:
\begin{align*}
z^{\pi^{\nu_1}_1} + z^{\pi^{\nu_2}_1} + z^{\pi^{\nu_3}_1} + z^{\pi^{\nu_1}_2} + z^{\pi^{\nu_2}_2} + (1-z^{\pi^{\nu_3}_2}) \geq 1   
\end{align*}
which removes all the  infeasible binary sequences $[0, 0, 0; 1, 0, 0; *, *, *]$, where $*$ can take arbitrary values. 
\end{example}

Furthermore, by leveraging our special optimization formulation, where the BMP schedules PoIs visit timings and sequence and the BSP generates trajectories from one PoI to the next, the time-shifted cuts take a simpler form. If the robot departs from $\mathcal{R}_{\nu_i}$ at time $k_i$ and cannot reach $\mathcal{R}_{\nu_j}$ at time $k_j$ within $\Delta k = k_j - k_i$, then there is no way it can reach the same PoI within a shorter timespan, regardless of the starting time. In particular, if the BSP reports infeasibility to travel from $\mathcal{R}_{\nu_i}$ to $\mathcal{R}_{\nu_j}$ within $\Delta k$ steps, we add the following cuts:
\begin{align}
\forall \tau \in &\{0, \ldots, \Delta k\}, \; \forall k \in \{1, \ldots, K-\tau\}: \nonumber \\
&(1 - z^{\pi_k^{\nu_i}}) + (1 - z^{\pi_{k+\tau}^{\nu_j}}) \geq 1
\label{eqn:shifted_cut1}
\end{align}
%
Each cut eliminates all the task schedules that contain the segment in which the robot first visits $\mathcal{R}_{\nu_i}$ at step $k$ and then arrives at $\mathcal{R}_{\nu_j}$ at step $k+\tau$, with $\tau < \Delta k$, for any beginning step $k$. These cuts assume that infeasibility arises solely from the reachability between one PoI and another, overlooking any constraint violations that may occur during the intermediate steps of the trajectory segment. While this assumption may not be general, it is consistent with our problem formulation where the BSP decomposes motion planning into independent point-to-point segments between consecutive PoIs.

\begin{example}
[continued] Again consider a scenario where the robot starts at PoI 1 and targets PoI 3. Suppose that the infeasible sequence is $\widehat{\boldsymbol{z}}^{\pi^\nu} = [1, 0, 0; *, *, *; 0, 0, 1]$, where the subproblem reports that the robot cannot travel from PoI 1 to PoI 3 within $\Delta k = 2$ steps. We derive that the following schedules are also infeasible: $[1, 0, 0; 0, 0, 1; *, *, *]$ and $[*, *, *; 1, 0, 0; 0, 0, 1]$, since attempting the same transition within $\Delta k = 1$ step is impossible. Following~\eqref{eqn:shifted_cut1}, we  generate two additional cuts along with the original cut:
\begin{align*}
(1 - z^{\pi_1^{\nu_1}}) + (1 - z^{\pi_2^{\nu_3}}) \geq 1, \quad 
(1 - z^{\pi_2^{\nu_1}}) + (1 - z^{\pi_3^{\nu_3}}) \geq 1
\end{align*}
\end{example}

\begin{figure*}[t!]
    \centering
    \includegraphics[width=1\textwidth]{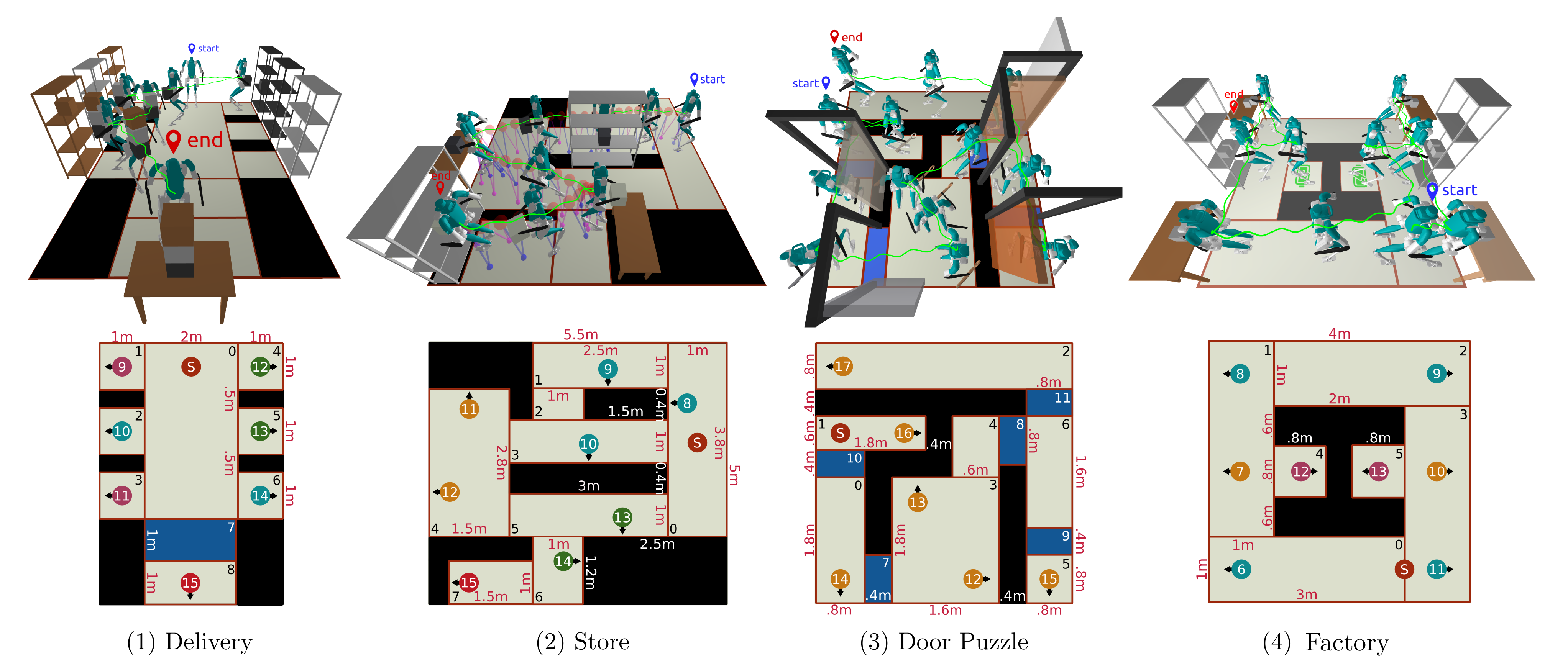}
     \caption{\textit{Top:} Time lapse for bipedal robot Digit to accomplish four task specifications in each environment. \textit{Bottom:} The layout and dimensions of each environment  are shown, where black regions indicate obstacles, dark red outlines represent convex regions, and colored dots denote PoIs with arrows indicating their heading angles. Each region and PoI is assigned an index to be used in the task specification. The robot starts from the red dot marked with ``S". The full recording of each solution can be viewed \href{https://bipedal-stl.github.io/}{here}.}
\label{fig:result}
\end{figure*} 

\section {Experiments and Results}

\begin{table}[t!]
\centering
\renewcommand{\arraystretch}{1.2}
\caption{List of preset values of parameters in the experiments} 
\label{Tab:parameters}
\begin{tabular}{|>{\centering\arraybackslash}m{0.8cm}|
>{\centering\arraybackslash}m{0.8cm}||
>{\centering\arraybackslash}m{0.8cm}|
>{\centering\arraybackslash}m{0.8cm}||
>{\centering\arraybackslash}m{0.8cm}|
>{\centering\arraybackslash}m{0.8cm}|}
\hline
$T$ & $0.4\,s$ & $H$ & $0.95\,m$ & $l^{\rm des}_k$ & $0.13\,m$ \\ \hline
$v_{\rm max}$ & $0.5\,m/s$ & $r_{\rm max}$ & $0.5\,m$ & $\alpha$ & $200$ \\ \hline
$\omega_2/\omega_1$ & $33.3$ & $\omega_3/\omega_1$ & $0.4$ & $\epsilon$ & $0.1$ \\ \hline
\end{tabular}
\begin{tabular}{|>{\centering\arraybackslash}m{1cm}|
>{\centering\arraybackslash}m{1.88cm}||
>{\centering\arraybackslash}m{1cm}|
>{\centering\arraybackslash}m{1.88cm}|}
$\boldsymbol{lb}$ & $[-0.2\,m; 0.2\,m]$ & $\boldsymbol{ub}$ & $[0.2\,m; 0.2\,m]$ \\ \hline
\end{tabular}
\end{table}
To demonstrate the efficiency of our proposed method using Benders Decomposition, we benchmark it against other formulations through several experiments with different task specifications. All our experiments run on a computer with a 13th Gen Intel Core i9-13900K CPU and 16 GB of memory. All MIPs involved in these methods are solved using the commercial solver Gurobi 12.0.

\subsection{Baseline Approaches}
\label{sec:experiment_baseline}

We compare our approach with two baseline MIP optimization formulations for solving bipedal TAMP problems involving STL constraints, reduced-order dynamics model, and nonlinear kinematic constraints, as follows:
\begin{itemize}
    \item \textbf{MIQCQP}: We implement a relaxed approach based on \cite{deits2014footstep} that utilizes the mixed-integer quadratically constrained quadratic program (MIQCQP) formulation for safe bipedal footstep planning, but with the goal removed from the objective function and instead incorporating STL constraints following the formulation 15 in \cite{belta2019formal}. In addition, we enforce foothold reachability constraints through an intersection of circular regions inscribed in the rectangular bounds of \eqref{eqn:kinematics}. The sine and cosine functions in the rotation matrix are approximated by piecewise linear functions that divides the robot's heading angle $(-\pi, \pi]$ into 4 equal-length intervals. To achieve obstacle avoidance, we decompose the configuration space into convex collision-free regions similar to what we do in the BD approach, as described in Sec.~\ref{sec:problem_formulation}. This baseline method solves the entire problem as a single MIQCQP without decomposition.
    \item \textbf{ADMM}: As another baseline for comparison, we solve the MINLP in Sec. \ref{sec:problem_formulation} using the well-established decomposition approach ADMM. Our ADMM implementation follows \cite{shirai2022simultaneous} and decomposes the problem into an MICP subproblem and an NLP projection step. The ADMM decomposition is similar to that of BD, with the key difference illustrated in Fig.~\ref{fig:admm_bd_comparison}. The constraints in the MICP and NLP subproblems are identical to those in our BMP and BSP, respectively, except that in the NLP, the task schedule $\boldsymbol{z}^{\pi^\nu}$ is not fixed. Instead, ADMM subproblems promote convergence through additional consensus terms in their objective functions that penalize deviations between residual variables (i.e., $\mathbf {X}$, $\mathbf {U}$, and $\mathbf {Z}$) from other subproblems using L2 norms weighted by a penalty parameter. The algorithm alternates between the MICP and NLP steps and updates the dual variables until stopping criteria, such as maximum residual tolerance or maximum iteration count, are met. We refer readers to Appendix~\ref{formulation_ADMM} for a detailed description of the algorithm.
\end{itemize}

\subsection{Benchmark Scenarios}
Our benchmark environments cover a range of application scenarios in which the robot is tasked to access a number of PoIs subject to certain logical and temporal conditions. Some of these environments are adapted from prior works \cite{kurtz2022mixed, sun2022multi}, but we extend the planning horizon to accommodate finer time discretization for footsteps and more complex task specifications. The corresponding task descriptions and specifications for each environment are listed below:
\begin{itemize}
    \item \textbf{Delivery} (Fig.~\ref{fig:result}(1)): The robot must collect a box from one shelf of each color before it can pass through Region 7 and ultimately reach the table to unload at Point 15. The task specification is $\varphi = \neg \pi^{7} \ \mathcal U_{[0, 70]}\ (\pi^9 \vee \pi^{10}) \wedge \neg \pi^{7} \ \mathcal U_{[0, 70]}\ (\pi^{11} \vee \pi^{13}) \wedge \neg \pi^{7} \ \mathcal U_{[0, 70]}\ (\pi^{12} \vee \pi^{14}) \wedge \Diamond_{[0, 70]} \pi^{15}$ 
    \item \textbf{Store} (Fig.~\ref{fig:result}(2)): The robot is instructed to reach one of Points 8, 9, or 10 within 15 steps after the start, and then sequentially reach one of Points 11 or 12, followed by one of Points 13, 14, and finally Point 15. The task specification is $\varphi = \Diamond_{[0, 15]} (\pi^8 \vee \pi^9 \vee \pi^{10}) \wedge \neg (\pi^{13} \vee \pi^{14}) \ \mathcal U_{[10, 50]}\ (\pi^{11} \vee \pi^{12}) \wedge \neg \pi^{15} \ \mathcal U_{[40, 70]}\ (\pi^{13} \vee \pi^{14}) \wedge \Diamond_{[50, 70]} \pi^{15}$.
    \item \textbf{Door Puzzle} (Fig.~\ref{fig:result}(3)): The robot is required to reach Point 17 at the end which can only be accessed by passing through several doors. Each door is paired with a key that must be collected from a designated PoI. The door-key pairings are outlined in the task specification: $\varphi = \neg \pi^{7} \ \mathcal U_{[0, 130]}\ \pi^{14}  \wedge \neg \pi^{8} \ \mathcal U_{[0, 130]}\ \pi^{12} \wedge \neg \pi^{9} \ \mathcal U_{[0, 130]}\ \pi^{13} \wedge \neg \pi^{11} \ \mathcal U_{[0, 130]}\ \pi^{15} \wedge \neg \pi^{10} \ \mathcal U_{[0, 130]}\ \pi^{16} \wedge \Diamond_{[0, 130]} \pi^{17}$.
    \item \textbf{Factory} (Fig.~\ref{fig:result}(4)): The robot must visit the charging stations at Points 12 and 13 within 50 steps each time it leaves a charging station. Additionally, after retrieving a box from one of the tables at Points 6, 8, 9, or 11, the robot must visit either Point 7 or 10 within 20 steps to unload it onto the shelves. The task specification is $\varphi = \square_{[0, 130]} \left(\neg (\pi^{12} \vee \pi^{13})\Rightarrow \Diamond_{[0, 50]}(\pi^{12} \vee \pi^{13})\right) \wedge \square_{[0, 130]} \left((\pi^{6}\vee \pi^{8}\vee \pi^{9}\vee \pi^{11}) \Rightarrow\Diamond_{[0, 20]}(\pi^{7} \vee \pi^{10})\right) \wedge \Diamond_{[0, 130]}\pi^6 \wedge \Diamond_{[0, 130]}\pi^8 \wedge \Diamond_{[0, 130]}\pi^9 \wedge \Diamond_{[0, 130]}\pi^{10}$.
\end{itemize}
The layout of each environment is presented in Fig. \ref{fig:result}. Each red box represents a convex region that partitions the workspace, and each dot represents a PoI. The arrows indicate the corresponding heading angle associated with each PoI. We assume that all obstacles are polyhedral and that the convex regions are rectangular, although this framework can be extended to handle overlapping convex polygons of arbitrary shapes. The LIP model used in these experiments is a simplified representation of the real bipedal robot, Digit. All the parameters in the constraints are set to the values in Table \ref{Tab:parameters}.

\begin{table}[t!]
\centering
\renewcommand{\arraystretch}{1.2}
\caption{List of data from the experiments in BD} 
\label{Tab:results2}
\begin{tabular}{|>{\centering\arraybackslash}m{3cm}|*4{>{\centering\arraybackslash}m{0.9cm}}|}
\hline
& Delivery & Store & Door Puzzle & Factory   \\ \hline \hline
No. of Binary Variables & 1775 & 1775 & 3537 & 3013 \\ \hline 
No. of Cont. Variables & 1476 & 1657 & 2545 & 2300 \\ \hline 
Initial No. of Constraints & 15393 & 10610 & 50903 & 26335 \\ \hline
Total No. of Cuts & 1447 & 2515 & 1830 & 7283 \\ \hline
BMP Solving Time Ratio  & 0.93 & 0.91 & 0.93 & 0.99 \\ \hline

Feas. Iteration  & 3  & 7  & 2 & 2  \\ \hline

Opt. Iteration  & 7  & 7  & 9  & 16 \\ \hline 



\end{tabular}
\end{table}

\begin{table}[t!]
\centering
\begin{threeparttable}
\renewcommand{\arraystretch}{1.2}
\caption{Comparison of runtime (seconds) to find the first feasible solution (Feas.) and the optimal solution (Opt.)*} 
\label{Tab:results1}
\begin{tabular}{|>{\centering\arraybackslash}m{1.2cm}|l|*3{>{\centering\arraybackslash}m{1.2cm}}|}
\hline
\multicolumn{2}{|c|}{} &
MIQCQP & ADMM& \textbf{BD}\\ \hline \hline
\multirow{2}{=}{Delivery} & Feas. & 3813  & 1336 & 26   \\ \cline{2-5}
           & Opt.  & 3813  & \scriptsize{N/A} & 56   \\ \hline \hline

\multirow{2}{=}{Store}     & Feas. & 6122  & 682 & 30   \\ \cline{2-5}
          & Opt.  & 9245  & \scriptsize{N/A} & 30   \\ \hline \hline

\multirow{2}{=}{Door Puzzle}  & Feas. &  \scriptsize{T.O.}  &  \scriptsize{T.O.} & 120   \\ \cline{2-5}
     & Opt.  & \scriptsize{T.O.}  & \scriptsize{N/A} & 329   \\ \hline \hline

\multirow{2}{=}{Factory}  & Feas.  & \scriptsize{T.O.} & 8596 & 56   \\ \cline{2-5}
     & Opt.  & \scriptsize{T.O.}  & \scriptsize{N/A} & 4006   \\ \hline
\end{tabular}
\begin{tablenotes}
  \scriptsize
  \item * T.O. indicates a time out when the computation time exceeds 10,000 seconds. Since ADMM is unable to find the optimal solution, all corresponding entries are reported as N/A.
\end{tablenotes}
\end{threeparttable}
\end{table}

\begin{table}[t!]
\centering
\renewcommand{\arraystretch}{1.2}
\caption{Comparison of total footsteps from the solutions \\ in BD and ADMM} 
\label{Tab:results3}
\begin{tabular}{|>{\centering\arraybackslash}m{1.5cm}|*4{>{\centering\arraybackslash}m{0.9cm}}|}
\hline
& Delivery & Store & Door Puzzle & Factory   \\ \hline\hline 

BD Feas. & 69  & 52    & 128  & 128 \\ \hline

BD Opt.  & 54  & 52    & 96  & 101 \\ \hline \hline

ADMM  & 58  & 57    & \scriptsize{T.O}  & 115 \\ \hline
\end{tabular}
\end{table}

\begin{table}[ht]
    \begin{minipage}[c]{0.4\linewidth}
        \centering        \includegraphics[width=25mm]{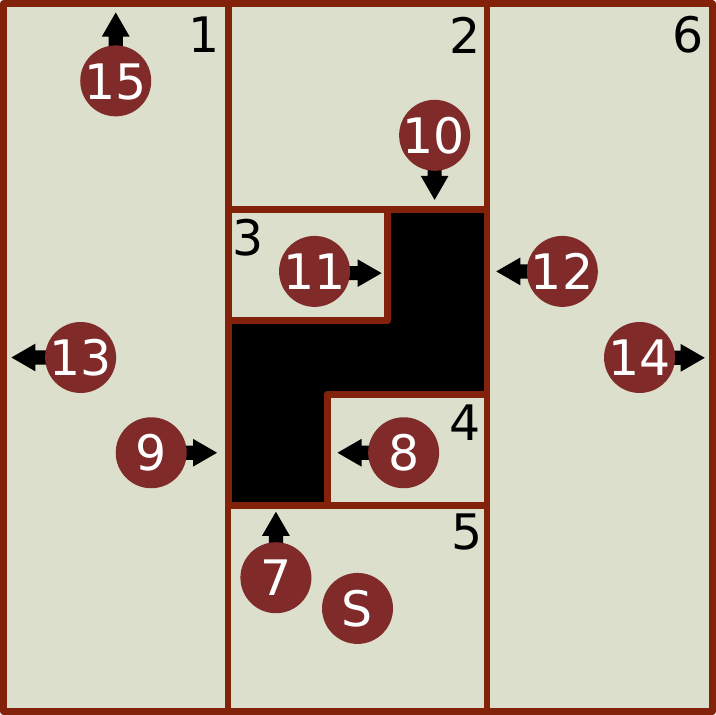}
        \vspace{-0.5em}
        \captionof{figure}{The map for abalation study of the time-shifted cuts.}
        \label{fig:abalation}
  \end{minipage}
  \hfill
  \begin{minipage}[c]{0.6\linewidth}
    \vspace{1.5em}
    \centering
    \renewcommand{\arraystretch}{1.2}
    \begin{tabular}{|l|c c|c c|}
        \hline
        & \multicolumn{2}{c|}{Feas.} & \multicolumn{2}{c|}{Opt.} \\ \cline{2-5}
        & w/ & w/o & w/ & w/o \\ \hline\hline
        Iteration & 3 & 10 & 10 & 25 \\ \hline
        Runtime (s) & 20 & 74 & 65 & 264 \\ \hline
    \end{tabular}
    \vspace{1em}
    \caption{Comparison of iteration and runtime with and without time-shifted cuts}
    \label{table:abalation}
  \end{minipage}
\end{table}

\subsection{Results and Observations}

The optimal CoM trajectory and the robot's motion generated by our proposed approach are shown in Fig. \ref{fig:result}. In Store environment, we also visualize the corresponding LIP model at each foot contact instant. In Table \ref{Tab:results3}, we record the computation time to find the first feasible and the optimal solutions across the four benchmark scenarios for our proposed BD method, comparing against the baseline MIQCQP and ADMM methods. In Table \ref{Tab:results2}, we report the details of the BD formulation and computation performance, including the number of binary and continuous variables, constraints, the total number of iterations required to find the feasible and optimal solutions, the total number of feasibility and optimality cuts added to the BMP during these iterations, and the relative proportion of BMP solving time to total solving time. We also record the number of footsteps to complete the entire task using the first feasible and optimal solutions obtained from the BD method, and compare these with the results from the baseline ADMM method. 

Results in Table~\ref{Tab:results1} show that our proposed BD approach achieves a speedup of more than an order of magnitude over the MIQCQP and ADMM baselines, while still maintaining the solution optimality. In comparison, the non-decomposed MIQCQP method often requires several thousand seconds to find a feasible solution and frequently times out for longer planning horizons due to the large number of binary variables introduced by the STL constraints. The BD approach finds a feasible solution in tens of seconds and reaches the optimal solution with only a modest increase in computation time, except in Factory environment, where the baseline methods either time out or fail to find an optimal solution. The longer computation time in Factory environment to find an optimal solution arises from more candidate schedules, which results from the additional requirement to visit the charging station in a timely manner after performing other tasks for a certain duration. 



\begin{figure}[t]
\centering
\includegraphics[width=\columnwidth]{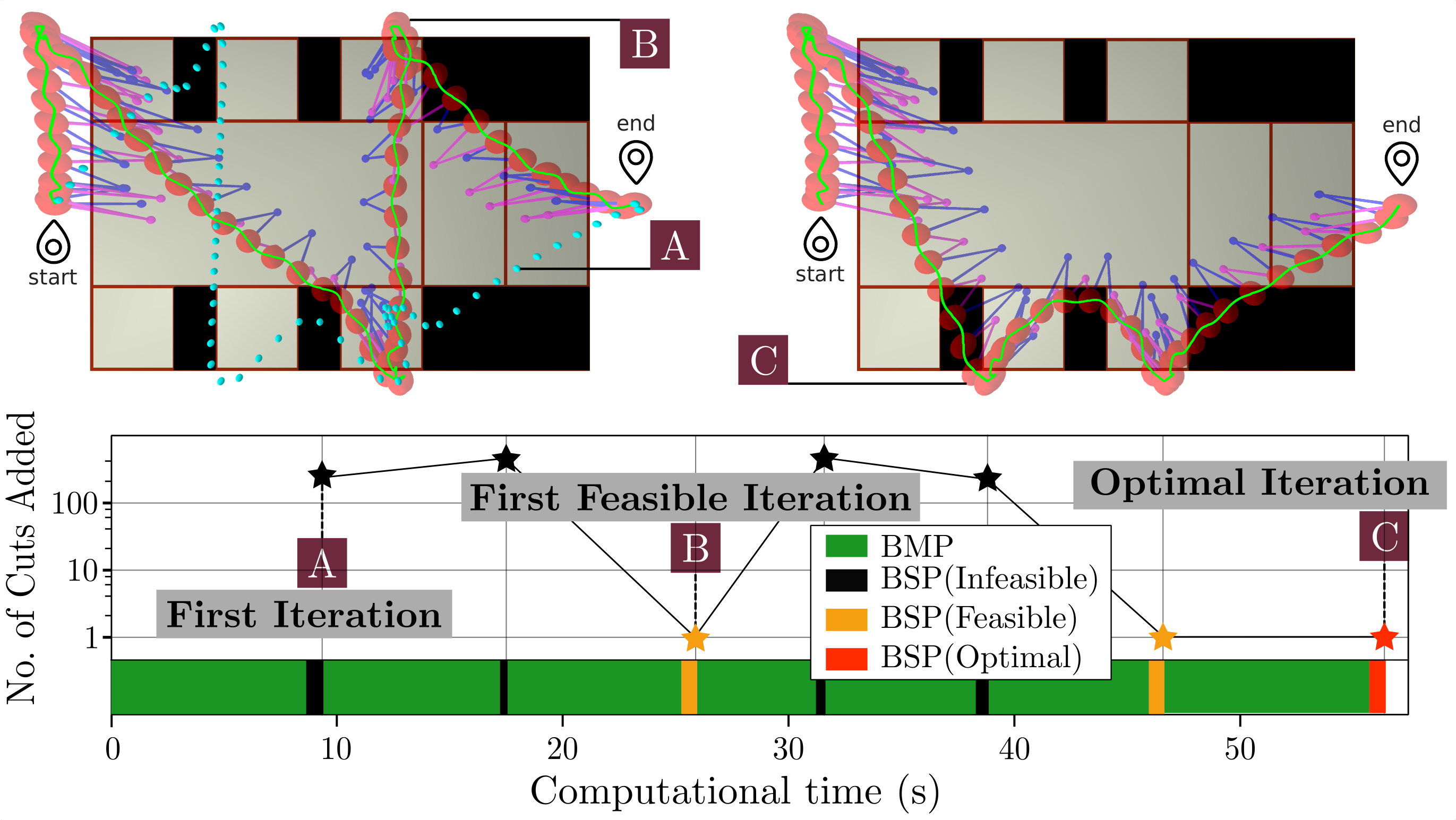}
    \caption{Iterations of solutions generated by our proposed method in Delivery environment. The stripe at the bottom shows the time spent by the BMP and the BSPs, along with the number of cuts added to the BMP at the end of each iteration. The black stars indicate moments when the BSPs are solved as infeasible and multiple feasibility cuts are added, while the yellow stars represent iterations when all BSPs are solved as feasible and only an optimality cut is added. We highlight three key moments: an infeasible solution (A) which is shown as the cyan dots on the top left figure, the first feasible solution (B) which is shown by a sequence of LIP models at each contact instant on the top left figure, and the optimal solution (C) that is represented by a sequence of LIP models at each contact instant on the top right figure.}
    \label{fig:iterations}
\end{figure}

Our time-shifted cuts generation strategy, introduced in Sec.~\ref{sec:cutting plane}, plays a critical role in accelerating the discovery of both feasible and optimal solutions. This is evidenced by an ablation test in which only a single cut is added using constraint~\eqref{eqn:feas_cut} after each iteration, instead of generating multiple time-shifted cuts at once. We observe that when only a single cut is added, no feasible solution is found within 100 iterations across all benchmark environments. Our proposed time-shifted cutting plane technique, however, consistently finds the optimal solution within 20 iterations, as shown in Table~\ref{Tab:results2}. In Fig.~\ref{fig:iterations}, we list the number of feasibility and optimality cuts added after each iteration in Delivery environment, and the evolution of the solutions after applying these cuts to the BMP. For further comparison, we present a scenario in Fig.~\ref{fig:abalation} where the robot can find a solution within a reasonable time even without time-shifted cuts. The task specifition is given by $\varphi = \left( \neg \pi^{13} \mathcal \ U_{[0, 70]} \ (\pi^{7} \vee \pi^{8} \vee \pi^{9}) \wedge \neg \pi^{15} \ \mathcal U_{[0, 70]} \ \pi^{13} \right) \vee \left( \neg \pi^{14} \ \mathcal U_{[0, 70]} \ (\pi^{10} \vee \pi^{11} \vee \pi^{12}) \wedge  \neg \pi^{15} \ \mathcal U_{[0, 70]} \ \pi^{14} \right) \wedge \Diamond_{[0, 70]} \pi^{15} $. As shown in Table~\ref{table:abalation}, applying the time-shifted cuts significantly shortens the time required to obtain the first feasible or optimal solution, demonstrating the crucial role of our novel cut-generation strategy in accelerating the solution finding.

Despite the strengths of our proposed BD approach,  the binary variables in the BMP still cause it to account for at least 90\% of the total computation time. This remains the primary factor limiting scalability and constrains the applicability of our approach to more complex environments and tasks that require multi-agent coordination. Future work could explore alternative dynamics simplification techniques, such as using Bézier curves to represent motion priors in the BMP, to further improve the overall solving speed of the framework.

\subsection{Comparison with ADMM}
Since ADMM employs a decomposition strategy similar to ours, we dedicate this subsection to a more detailed comparison between the two methods. We implement the ADMM formulation to solve the MINLP described in Sec.~\ref{sec:problem_formulation} following the approach in \cite{shirai2022simultaneous}, with two alternative strategies that modify the consensus terms and constraint implementations for the MICP and NLP subproblems. The complete ADMM formulation is provided in Appendix \ref{formulation_ADMM}. 


In Fig.~\ref{fig:ADMM_convergence}, we present the results from Store environment, which showcases one of the best convergence scenarios of ADMM. We show its convergence curves for the position residual $\max_k \|\boldsymbol{x}_k^{\text{MIP}} - \boldsymbol{x}_k^{\text{NLP}}\|_\infty$, direction residual $\max_k |\theta_k^{\text{MIP}} - \theta_k^{\text{NLP}}|$, and control input residual $\max_k \|\boldsymbol{u}_k^{\text{MIP}} - \boldsymbol{u}_k^{\text{NLP}}\|_\infty$. We also visualize the final solution obtained at the 23rd iteration after the convergence threshold is met, alongside BD's final optimal solution.
Similar to the observations in \cite{shirai2022simultaneous}, ADMM convergence is non-monotonic, and residuals can still increase even after many iterations. 
In most cases, residuals remain significant after tens of iterations despite large penalties for consensus violations. This indicates that the MICP solution from ADMM does not fully satisfy the foot kinematic reachability constraints~\eqref{eqn:kinematics}, and the NLP solution does not exactly satisfy the STL constraints~\eqref{eqn:connectives}. In comparison, all the feasible solutions that the BD method obtains strictly satisfy all 
the constraints.

As for the reason,  although ADMM guarantees optimal solutions for convex problems, it generally lacks such guarantees for the non-convex problems considered in this study. Given our objective of minimizing task completion time, ADMM often converges to suboptimal solutions that result in longer task durations (i.e., more total footsteps to complete all tasks), as shown in Table~\ref{Tab:results2}. 
The BD approach, on the other hand, systematically improves solution optimality by incorporating optimality cuts.

These observations highlight the fundamental differences between ADMM’s consensus-based strategy and the BD's propose-and-reject strategy. ADMM employs a consensus mechanism between variables, where the entire set of variables is exchanged between subproblems to guide convergence toward feasibility and optimality. This coordinated exchange is often more informative than the BD's iterative propose-and-reject mechanism where the BSP provides limited feedback to help the BMP adjust its decisions to find a feasible solution. As a result, ADMM converges more quickly to an approximate solution. However, this iterative interaction sometimes makes it difficult to guarantee exact convergence, especially for nonconvex problems.

On the other hand, BD's efficiency can be significantly enhanced through better cut generation strategies. In this study, we propose discovering multiple strong feasibility cuts in a single iteration by leveraging the structure of the BSP, without requiring additional computation to solve the BSP. As the abalation study in Fig.~\ref{fig:abalation} and Table~\ref{table:abalation} demonstrate, this cut generation strategy dramatically improves BD's convergence, allowing it to find solutions faster than ADMM. 

Finally, ADMM requires extensive tuning of penalty weights for each specific problem instance. Since every solving attempt takes considerable time, the parameter tuning process becomes expensive. In contrast, BD does not require tuning consensus penalty parameters or dual variable weights that are important for ADMM's convergence, making it easier to use for practitioners.

\begin{figure}[t]
\centering
\includegraphics[width=\columnwidth]{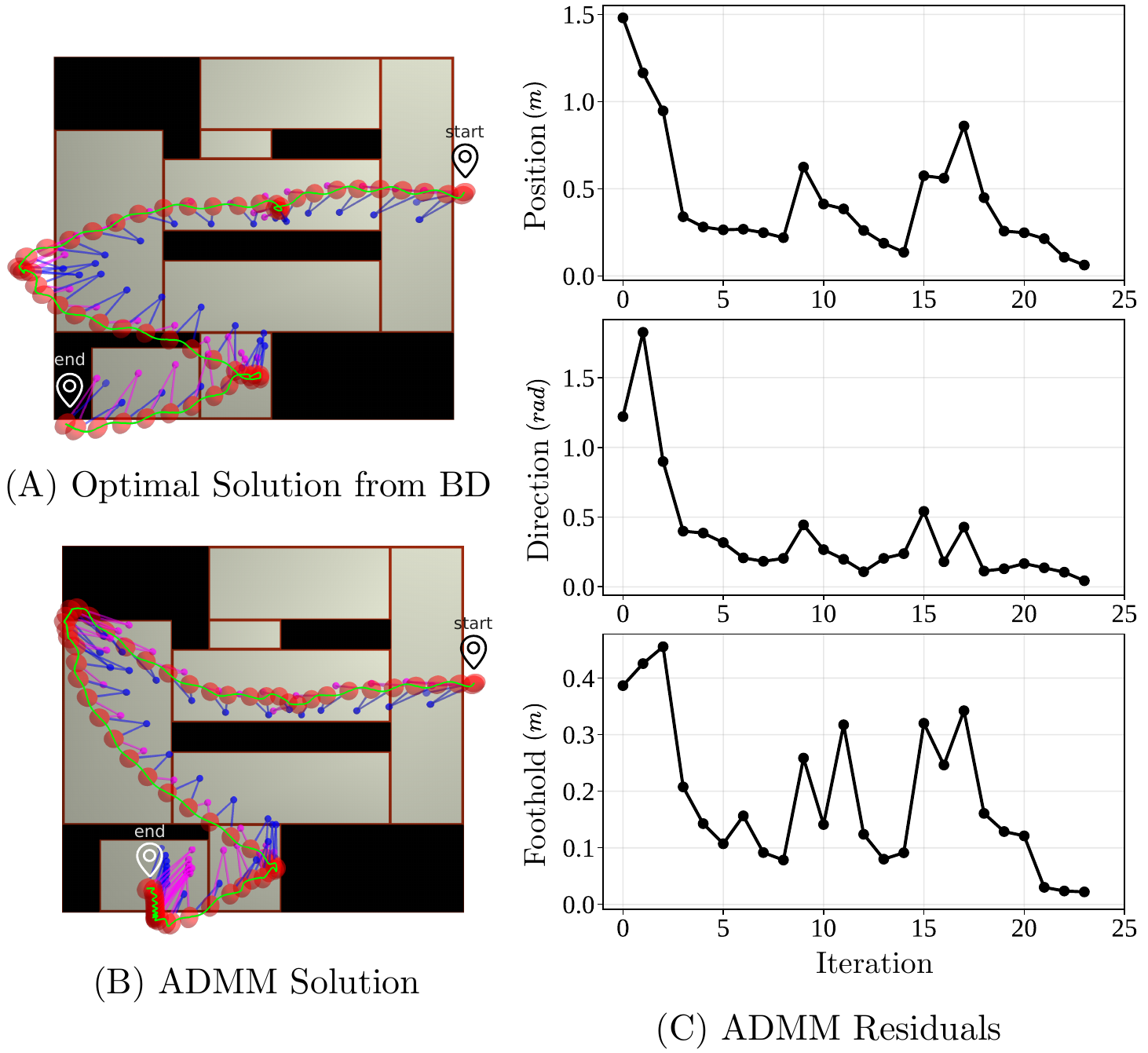}
    \caption{ADMM convergence for Store environment showing primal residuals of the position, direction and foothold over iterations. We also compare the result after ADMM hits the threshold and the optimal solution obtained from the BD method in (A) and (B). The Store environment is one of the best convergence scenarios for ADMM among all the studied environments. 
    }
    \label{fig:ADMM_convergence}
\end{figure} 

\section {Conclusion}
This work presents a novel distributed approach to accelerate the solving of bipedal TAMP problems under STL specifications using Benders Decomposition. Our approach significantly outperforms baseline methods and has the potential to be extended to other TAMP problems involving different types of robots with varying nonlinear dynamics models. The main bottleneck is that the master problem still accounts for a substantial portion of the total computation time. Future work will focus on improving the efficiency of solving the master problem to further enhance overall performance.

\begin{appendices}
\section{Cut Shifting Theorem and Proof}
\label{proof_cut_shifting}

In this appendix, we present a rigorous theorem for generating time-shifted cuts in the context of Benders Decomposition with BSP formulated as MIQP. The key insight is that when we identify an infeasible task schedule, we can discover multiple strong cutting planes that provably exclude other infeasible schedules without requiring additional BSP computations. These cuts leverage the temporal structure of the BSP: if reaching a target at time $k$ is infeasible, then reaching the same target at any earlier time $k' < k$ must be also infeasible under certain conditions.

Define an MIQP problem $\mathcal{P}(\boldsymbol{x}_{0}, \boldsymbol{\mu}, \boldsymbol{\nu})$ as:
\begin{subequations}
\begin{align}
\underset{\boldsymbol{x}, \boldsymbol{u}, \boldsymbol{\mu}, \boldsymbol{\nu}}{\text{minimize}} &\quad \sum_{k=0}^{K-1} \left( \boldsymbol{x}[k]^\top \boldsymbol{Q}_x \boldsymbol{x}[k] + \boldsymbol{u}[k]^\top \boldsymbol{Q}_u \boldsymbol{u}[k] \right) \nonumber \\
\text{s.t.} &\quad \boldsymbol{x}[0] = \boldsymbol{x}_0 \text{ (given)} \label{MIQP:initial_condition} \\
&\quad \text{For all } k \in \{0,\ldots,K-1\}: \nonumber \\
&\quad \boldsymbol{x}[k+1] = \boldsymbol{A}\boldsymbol{x}[k] + \boldsymbol{B}\boldsymbol{u}[k] + \boldsymbol{G}_{\mu}\boldsymbol{\mu}[k] \label{MIQP:con1}\\
&\quad \boldsymbol{C}\boldsymbol{x}[k] + \boldsymbol{D}\boldsymbol{u}[k] + \boldsymbol{H}_{\mu}\boldsymbol{\mu}[k] + \boldsymbol{H}_{\nu}\boldsymbol{\nu}[k] \leq \boldsymbol{h} \label{MIQP:con2} \\
&\quad \boldsymbol{\mu}[k] \in \{0,1\}^{n_{\mu}} \triangleq \mathcal{D}_{\mu}, \quad \boldsymbol{\nu}[k] \in \{0,1\}^{n_{\nu}} \triangleq \mathcal{D}_{\nu} \nonumber
\end{align}   
\end{subequations}
where $n_x$, $n_u$, $n_{\mu}$, $n_{\nu}$, and $n_c$ denote the dimensions of the state, control input, two sets of binary variables, and inequality constraints, respectively. $\boldsymbol{A} \in \mathbb{R}^{n_x \times n_x}$ is the state transition matrix, $\boldsymbol{B} \in \mathbb{R}^{n_x \times n_u}$ is the control input matrix, $\boldsymbol{G}_{\mu} \in \mathbb{R}^{n_x \times n_{\mu}}$ maps binary variables to state dynamics. $\boldsymbol{C} \in \mathbb{R}^{n_c \times n_x}$, $\boldsymbol{D} \in \mathbb{R}^{n_c \times n_u}$, $\boldsymbol{H}_{\mu} \in \mathbb{R}^{n_c \times n_{\mu}}$, and $\boldsymbol{H}_{\nu} \in \mathbb{R}^{n_c \times n_{\nu}}$ define the inequality constraints, and $\boldsymbol{h} \in \mathbb{R}^{n_c}$ is the inequality bounds.


In addition, our proof of the theorem relies on the following infeasibility certificates obtained from Farkas’ lemma (Section 5.8.3 in \cite{boyd2004convex}):

\begin{lemma}
\label{lem:farkas}
Given $\boldsymbol{E} \in \mathbb{R}^{l \times n}$, $\boldsymbol{e} \in \mathbb{R}^{l}$, $\boldsymbol{F} \in \mathbb{R}^{m \times n}$, $\boldsymbol{f} \in \mathbb{R}^{m}$, exactly one of the following statements is true:
\begin{enumerate}
    \item There exists an $\boldsymbol{x} \in \mathbb{R}^{n}$ that satisfies $\boldsymbol{E}\boldsymbol{x} = \boldsymbol{e}, \boldsymbol{F}\boldsymbol{x} \leq \boldsymbol{f}$.
    \item There exist $\boldsymbol{\pi} \in \mathbb{R}^{l}$, $\boldsymbol{\lambda} \in \mathbb{R}^{m}$ that satisfy $\boldsymbol{\lambda} \geq \boldsymbol{0}$, $\boldsymbol{E}^\top \boldsymbol{\pi} + \boldsymbol{F}^\top \boldsymbol{\lambda} = \boldsymbol{0}$, and $\boldsymbol{e}^\top \boldsymbol{\pi} + \boldsymbol{f}^\top \boldsymbol{\lambda} < 0$.
\end{enumerate}
\end{lemma}

We now present the following theorem for time-shifted cuts generation.
\begin{theorem}
\label{theorem2}
Consider the MIQP problem $\mathcal{P}(\boldsymbol{x}_{0}, \boldsymbol{\mu}, \widehat{\boldsymbol{\nu}})$, where $\widehat{\boldsymbol{\nu}} = \{\widehat{\boldsymbol{\nu}}[k]\}_{k=0}^{K-1}$ is a sequence that renders the problem infeasible for any $\boldsymbol{\mu} \in (\mathcal{D}_{\mu})^K$.  
If the following conditions hold:
\begin{enumerate}
    \item The initial state is stationary: $\boldsymbol{x}_0 = \boldsymbol{A}\boldsymbol{x}_0$
    \item Constraint \eqref{MIQP:con2} is satisfiable at $k=0$: $\exists \boldsymbol{u}[0]$ such that $\boldsymbol{C}\boldsymbol{x}_0 + \boldsymbol{D}\boldsymbol{u}[0] + \boldsymbol{H}_{\nu}\widehat{\boldsymbol{\nu}}[0] \leq \boldsymbol{h}$
    \item The matrices $\boldsymbol{B}^\top$ and $\boldsymbol{D}^\top$ have disjoint sets of non-zero rows, i.e., $\{i : \boldsymbol{B}^\top_{i,:} \neq \boldsymbol{0}\} \cap \{i : \boldsymbol{D}^\top_{i,:} \neq \boldsymbol{0}\} = \emptyset$
\end{enumerate}
then the time-shifted sequence $${}^s\widehat{\boldsymbol{\nu}} = \{\widehat{\boldsymbol{\nu}}[1], \widehat{\boldsymbol{\nu}}[2], \ldots, \widehat{\boldsymbol{\nu}}[K-1], \boldsymbol{*}\}$$ also renders problem $\mathcal{P}(\boldsymbol{x}_{0}, \boldsymbol{\mu}, {}^s\widehat{\boldsymbol{\nu}})$ infeasible for any $\boldsymbol{\mu} \in (\mathcal{D}_{\mu})^K$, where $\boldsymbol{*} \in \mathcal{D}_{\nu}$ can take arbitrary values.
\end{theorem}

\begin{proof}
For our MIQP problem, once $\boldsymbol{\mu}$ and $\boldsymbol{\nu}$ are fixed, the remaining program reduces to a linear system in $\boldsymbol{x}$ and $\boldsymbol{u}$. 
Farkas' theorem is applied to characterize the infeasibility of this system. The condition coincides with equation (16) in \cite{lin2024accelerate}, which states that the linear system formed by constraints \eqref{MIQP:initial_condition}, \eqref{MIQP:con1}, and \eqref{MIQP:con2} for all $k \in \{0, 1, \ldots, K-1\}$ is infeasible if and only if there exist certificates:
\begin{align}
\boldsymbol{\lambda} &= \{\boldsymbol{\lambda}[k]\}_{k=0}^{K-1} \\
\boldsymbol{\pi} &= \{\boldsymbol{\pi}[k]\}_{k=0}^{K}
\end{align}
that satisfy:
\allowdisplaybreaks
\begin{subequations}
\begin{align}
& \boldsymbol{\lambda}[k] \geq \boldsymbol{0}, \quad k = 0, 1, \ldots, K-1 \label{eq:farkas_a}\\
& \text{For } k = 0, 1, \ldots, K-1: \nonumber \\
& \quad \boldsymbol{A}^\top\boldsymbol{\pi}[k+1] = \boldsymbol{\pi}[k] + \boldsymbol{C}^\top\boldsymbol{\lambda}[k] \label{eq:farkas_b}\\
& \quad \boldsymbol{B}^\top \boldsymbol{\pi}[k+1] = \boldsymbol{D}^\top \boldsymbol{\lambda}[k] \label{eq:farkas_c} \\
& \boldsymbol{x}_0^\top\boldsymbol{\pi}[0] + \sum_{k=0}^{K-1}(\boldsymbol{G}_{\mu}\boldsymbol{\mu}[k])^\top\boldsymbol{\pi}[k+1] \nonumber \\
& \quad + \sum_{k=0}^{K-1}(\boldsymbol{h} - \boldsymbol{H}_{\mu}\boldsymbol{\mu}[k] - \boldsymbol{H}_{\nu}\boldsymbol{\nu}[k])^\top\boldsymbol{\lambda}[k] < 0 \label{eq:farkas_d}
\end{align}
\end{subequations}
%
Since $\widehat{\boldsymbol{\nu}}$, the inverse-shift of ${}^s\widehat{\boldsymbol{\nu}}$, leads to infeasibility, we build Farkas certificates for an inverse-shifted version of $\boldsymbol{\mu}$, denoted  ${}^i\boldsymbol{\mu}$. We then forward-shift these certificates to obtain valid Farkas certificates for the original $\boldsymbol{\mu}$ under ${}^s\widehat{\boldsymbol{\nu}}$. Define ${}^i\boldsymbol{\mu}$ as:
$${}^i\boldsymbol{\mu} = \{\boldsymbol{0}, \boldsymbol{\mu}[0], \boldsymbol{\mu}[1], \ldots, \boldsymbol{\mu}[K-2]\}$$
Since ${}^i\boldsymbol{\mu} \in (\mathcal{D}_{\mu})^K$, the problem $\mathcal{P}(\boldsymbol{x}_0, {}^i\boldsymbol{\mu}, \widehat{\boldsymbol{\nu}})$ is also infeasible. By Farkas' theorem, there exist certificates:
\begin{align}
\boldsymbol{\lambda}_{({}^i\boldsymbol{\mu})} &= \{\boldsymbol{\lambda}_{({}^i\boldsymbol{\mu})}[k]\}_{k=0}^{K-1} \\
\boldsymbol{\pi}_{({}^i\boldsymbol{\mu})} &= \{\boldsymbol{\pi}_{({}^i\boldsymbol{\mu})}[k]\}_{k=0}^{K}
\end{align}
that satisfy conditions \eqref{eq:farkas_a}--\eqref{eq:farkas_d} with $\boldsymbol{\lambda}_{({}^i\boldsymbol{\mu})}$, $\boldsymbol{\pi}_{({}^i\boldsymbol{\mu})}$, ${}^i\boldsymbol{\mu}$, and $\widehat{\boldsymbol{\nu}}$ in place of $\boldsymbol{\lambda}$, $\boldsymbol{\pi}$, $\boldsymbol{\mu}$, and $\boldsymbol{\nu}$, respectively.

Define the shifted Farkas certificates $\boldsymbol{\lambda}_{({}^i\boldsymbol{\mu})}$, $\boldsymbol{\pi}_{({}^i\boldsymbol{\mu})}$ as:
\begin{align}
{}^s\boldsymbol{\lambda}_{({}^i\boldsymbol{\mu})} &= \{\boldsymbol{\lambda}_{({}^i\boldsymbol{\mu})}[1], \ldots, \boldsymbol{\lambda}_{({}^i\boldsymbol{\mu})}[K-1], \boldsymbol{0}\} \\
{}^s\boldsymbol{\pi}_{({}^i\boldsymbol{\mu})} &= \{\boldsymbol{\pi}_{({}^i\boldsymbol{\mu})}[1], \ldots, \boldsymbol{\pi}_{({}^i\boldsymbol{\mu})}[K-1], \boldsymbol{\pi}_{({}^i\boldsymbol{\mu})}[K], \boldsymbol{0}\}
\end{align}

To prove that the subproblem $\mathcal{P}(\boldsymbol{x}_0, \boldsymbol{\mu}, {}^s\widehat{\boldsymbol{\nu}})$ is infeasible for any $\boldsymbol{\mu}$, we claim that by replacing $\boldsymbol{\lambda}_{({}^i\boldsymbol{\mu})}$, $\boldsymbol{\pi}_{({}^i\boldsymbol{\mu})}$, ${}^i\boldsymbol{\mu}$, and $\widehat{\boldsymbol{\nu}}$ with ${}^s\boldsymbol{\lambda}_{({}^i\boldsymbol{\mu})}$, ${}^s\boldsymbol{\pi}_{({}^i\boldsymbol{\mu})}$, $\boldsymbol{\mu}$,  and ${}^s\widehat{\boldsymbol{\nu}}$, conditions \eqref{eq:farkas_a}--\eqref{eq:farkas_d} are still satisfied. ${}^s\boldsymbol{\lambda}_{({}^i\boldsymbol{\mu})}$ and ${}^s\boldsymbol{\pi}_{({}^i\boldsymbol{\mu})}$ thereby serve as the Farkas certificates for the infeasibility of $\mathcal{P}(\boldsymbol{x}_0, \boldsymbol{\mu}, {}^s\widehat{\boldsymbol{\nu}})$.

We now establish the property $\boldsymbol{\lambda}_{({}^i\boldsymbol{\mu})}[0] = \boldsymbol{0}$. Suppose that a Farkas certificate exists with $\boldsymbol{\lambda}_{({}^i\boldsymbol{\mu})}[0] \neq \boldsymbol{0}$. In that case, one can construct an alternative Farkas certificate in which $\boldsymbol{\lambda}'_{({}^i\boldsymbol{\mu})}[0] = \boldsymbol{0}$, effectively attributing the infeasibility to the dynamic constraints rather than to the inequality constraint at $k=0$. Formally, given the certificates $(\boldsymbol{\pi}_{({}^i\boldsymbol{\mu})}, \boldsymbol{\lambda}_{({}^i\boldsymbol{\mu})})$ with $\boldsymbol{\lambda}_{({}^i\boldsymbol{\mu})}[0] \neq \boldsymbol{0}$, we let:
\begin{align*}
\boldsymbol{\lambda}'_{({}^i\boldsymbol{\mu})}[k] &= \begin{cases}
\boldsymbol{0} & \text{if } k = 0 \\
\boldsymbol{\lambda}_{({}^i\boldsymbol{\mu})}[k] & \text{if } k = 1, \ldots, K-1
\end{cases} \\
\boldsymbol{\pi}'_{({}^i\boldsymbol{\mu})}[k] &= \begin{cases}
\boldsymbol{\pi}_{({}^i\boldsymbol{\mu})}[0] + \boldsymbol{C}^\top \boldsymbol{\lambda}_{({}^i\boldsymbol{\mu})}[0] & \text{if } k = 0 \\
\boldsymbol{\pi}_{({}^i\boldsymbol{\mu})}[k] & \text{if } k = 1, \ldots, K
\end{cases}
\end{align*}
This modified construction satisfies all Farkas conditions \eqref{eq:farkas_a}--\eqref{eq:farkas_d}. Condition \eqref{eq:farkas_a} holds trivially. For $k = 0$, condition \eqref{eq:farkas_b} becomes $\boldsymbol{A}^\top \boldsymbol{\pi}'_{({}^i\boldsymbol{\mu})}[1] = \boldsymbol{A}^\top \boldsymbol{\pi}_{({}^i\boldsymbol{\mu})}[1] = \boldsymbol{\pi}_{({}^i\boldsymbol{\mu})}[0] + \boldsymbol{C}^\top \boldsymbol{\lambda}_{({}^i\boldsymbol{\mu})}[0] = \boldsymbol{\pi}'_{({}^i\boldsymbol{\mu})}[0]$ and is therefore satisfied. For $k \geq 1$, the condition is unchanged and thus remains valid.

For condition \eqref{eq:farkas_c} at $k = 0$, we need to verify $\boldsymbol{B}^\top \boldsymbol{\pi}'_{({}^i\boldsymbol{\mu})}[1] = \boldsymbol{D}^\top \boldsymbol{\lambda}'_{({}^i\boldsymbol{\mu})}[0]$. By condition 3, $\boldsymbol{B}^\top$ and $\boldsymbol{D}^\top$ have disjoint sets of non-zero rows; that is, for each row index $i$, at least one of $\boldsymbol{B}^\top_{i,:}$ or $\boldsymbol{D}^\top_{i,:}$ is identically zero. Therefore, the equality $\boldsymbol{B}^\top \boldsymbol{\pi}_{({}^i\boldsymbol{\mu})}[1] = \boldsymbol{D}^\top \boldsymbol{\lambda}_{({}^i\boldsymbol{\mu})}[0]$ can hold only if both sides are zero in every row. This gives us $\boldsymbol{B}^\top \boldsymbol{\pi}'_{({}^i\boldsymbol{\mu})}[1] = \boldsymbol{B}^\top \boldsymbol{\pi}_{({}^i\boldsymbol{\mu})}[1] = \boldsymbol{D}^\top \boldsymbol{\lambda}'_{({}^i\boldsymbol{\mu})}[0] = \boldsymbol{0}$, which proves condition \eqref{eq:farkas_c} at $k = 0$. For $k \geq 1$, this condition continues to hold.

Finally, we show that condition \eqref{eq:farkas_d} is true. The left-hand side (LHS) with new certificates is:
\begin{align*}
&\boldsymbol{x}_0^\top \boldsymbol{\pi}'_{({}^i\boldsymbol{\mu})}[0] + \sum_{k=0}^{K-1}(\boldsymbol{G}_{\mu}{}^i\boldsymbol{\mu}[k])^\top\boldsymbol{\pi}'_{({}^i\boldsymbol{\mu})}[k+1] \nonumber \\
&\quad + \sum_{k=0}^{K-1}(\boldsymbol{h} - \boldsymbol{H}_{\mu}{}^i\boldsymbol{\mu}[k] - \boldsymbol{H}_{\nu}\widehat{\boldsymbol{\nu}}[k])^\top\boldsymbol{\lambda}'_{({}^i\boldsymbol{\mu})}[k] \\
&= \boldsymbol{x}_0^\top (\boldsymbol{\pi}_{({}^i\boldsymbol{\mu})}[0] + \boldsymbol{C}^\top \boldsymbol{\lambda}_{({}^i\boldsymbol{\mu})}[0]) \nonumber \\
&\quad + \sum_{k=0}^{K-1}(\boldsymbol{G}_{\mu}{}^i\boldsymbol{\mu}[k])^\top\boldsymbol{\pi}_{({}^i\boldsymbol{\mu})}[k+1] \nonumber \\
&\quad + \sum_{k=1}^{K-1}(\boldsymbol{h} - \boldsymbol{H}_{\mu}{}^i\boldsymbol{\mu}[k] - \boldsymbol{H}_{\nu}\widehat{\boldsymbol{\nu}}[k])^\top\boldsymbol{\lambda}_{({}^i\boldsymbol{\mu})}[k] \\
&= \text{(original LHS)} + \boldsymbol{x}_0^\top \boldsymbol{C}^\top \boldsymbol{\lambda}_{({}^i\boldsymbol{\mu})}[0] \nonumber \\
&\quad - (\boldsymbol{h} - \boldsymbol{H}_{\mu}{}^i\boldsymbol{\mu}[0] - \boldsymbol{H}_{\nu}\widehat{\boldsymbol{\nu}}[0])^\top \boldsymbol{\lambda}_{({}^i\boldsymbol{\mu})}[0]
\end{align*}

With ${}^i\boldsymbol{\mu}[0] = \boldsymbol{0}$, this becomes:
\begin{align*}
&= \text{(original LHS)} + (\boldsymbol{C}\boldsymbol{x}_0 + \boldsymbol{H}_{\nu}\widehat{\boldsymbol{\nu}}[0] - \boldsymbol{h})^\top \boldsymbol{\lambda}_{({}^i\boldsymbol{\mu})}[0]
\end{align*}

Since $\boldsymbol{D}^\top\boldsymbol{\lambda}_{({}^i\boldsymbol{\mu})}[0] = \boldsymbol{0}$ is established from condition 3, we can multiply both sides of the inequality from condition 2 (i.e., $\boldsymbol{C}\boldsymbol{x}_0 + \boldsymbol{D}\boldsymbol{u}[0] + \boldsymbol{H}_{\nu}\widehat{\boldsymbol{\nu}}[0] \leq \boldsymbol{h}$) by $\boldsymbol{\lambda}_{({}^i\boldsymbol{\mu})}[0] \geq \boldsymbol{0}$ to obtain $(\boldsymbol{C}\boldsymbol{x}_0 + \boldsymbol{H}_{\nu}\widehat{\boldsymbol{\nu}}[0] - \boldsymbol{h})^\top \boldsymbol{\lambda}_{({}^i\boldsymbol{\mu})}[0] \leq 0$. Because the original LHS is strictly negative due to the Farkas condition \eqref{eq:farkas_d}, adding a non-positive term preserves the strict negativity.

At this point, we list the key properties that establish the relationships between the original and shifted variables:
\begin{enumerate}[label=(\alph*)]
    \item ${}^s\boldsymbol{\pi}_{({}^i\boldsymbol{\mu})}[k] = \boldsymbol{\pi}_{({}^i\boldsymbol{\mu})}[k+1]$ for $k \in \{0,1,\ldots,K-1\}$ \\
    ${}^s\boldsymbol{\lambda}_{({}^i\boldsymbol{\mu})}[k] = \boldsymbol{\lambda}_{({}^i\boldsymbol{\mu})}[k+1]$ for $k \in \{0,1,\ldots,K-2\}$ \label{property:a}
    \item ${}^s\boldsymbol{\pi}_{({}^i\boldsymbol{\mu})}[K] = \boldsymbol{0}$ \\
    ${}^s\boldsymbol{\lambda}_{({}^i\boldsymbol{\mu})}[K-1] = \boldsymbol{0}$ \label{property:b}
    \item ${}^s\widehat{\boldsymbol{\nu}}[k] = \widehat{\boldsymbol{\nu}}[k+1]$ for $k \in \{0, 1, \ldots, K-2\}$ \label{property:c}
    \item ${}^i\boldsymbol{\mu}[0] = \boldsymbol{0}$, and ${}^i\boldsymbol{\mu}[k+1] = \boldsymbol{\mu}[k]$ for $k \in \{0, \ldots, K-2\}$\label{property:d}
    \item $\boldsymbol{\lambda}_{({}^i\boldsymbol{\mu})}[0] = \boldsymbol{0}$ \label{property:e}
    \item $\boldsymbol{x}_0^\top(\boldsymbol{\pi}_{({}^i\boldsymbol{\mu})}[1]) = \boldsymbol{x}_0^\top(\boldsymbol{\pi}_{({}^i\boldsymbol{\mu})}[0])$ \label{property:f}
\end{enumerate}


Properties \ref{property:a}--\ref{property:d} follow by construction. We have shown the property \ref{property:e}. Property \ref{property:f} follows from the stationarity condition (condition 1) together with property \ref{property:e}. In particular, \eqref{eq:farkas_b} implies $\boldsymbol{A}^\top\boldsymbol{\pi}_{({}^i\boldsymbol{\mu})}[1] = \boldsymbol{\pi}_{({}^i\boldsymbol{\mu})}[0] + \boldsymbol{C}^\top\boldsymbol{\lambda}_{({}^i\boldsymbol{\mu})}[0] \overset{\ref{property:e}}{=} \boldsymbol{\pi}_{({}^i\boldsymbol{\mu})}[0]$. Left-multiplying both sides by $\boldsymbol{x}_0^\top$, and using the stationarity condition $\boldsymbol{x}_0^\top\boldsymbol{A}^\top = \boldsymbol{x}_0^\top$, we obtain $\boldsymbol{x}_0^\top\boldsymbol{\pi}_{({}^i\boldsymbol{\mu})}[1] = \boldsymbol{x}_0^\top\boldsymbol{\pi}_{({}^i\boldsymbol{\mu})}[0]$, which is property (f).


With properties \ref{property:a}--\ref{property:d}, it is noticed that if we replace $\boldsymbol{\lambda}_{({}^i\boldsymbol{\mu})}$ and $\boldsymbol{\pi}_{({}^i\boldsymbol{\mu})}$ with ${}^s\boldsymbol{\lambda}_{({}^i\boldsymbol{\mu})}$ and ${}^s\boldsymbol{\pi}_{({}^i\boldsymbol{\mu})}$, conditions \eqref{eq:farkas_a}--\eqref{eq:farkas_c} are still satisfied.

It is left to verify condition \eqref{eq:farkas_d} for ${}^s\boldsymbol{\lambda}_{({}^i\boldsymbol{\mu})}$, ${}^s\boldsymbol{\pi}_{({}^i\boldsymbol{\mu})}$, $\boldsymbol{\mu}$ and ${}^s\widehat{\boldsymbol{\nu}}$. We start from the LHS of it and show that it remains negative through a series of transformations. At each step, we indicate which of the properties \ref{property:a}--\ref{property:f} established earlier is being used:
\allowdisplaybreaks
\begin{align}
&\boldsymbol{x}_0^\top({}^s\boldsymbol{\pi}_{({}^i\boldsymbol{\mu})}[0]) + \sum_{k=0}^{K-1}(\boldsymbol{G}_{\mu}\boldsymbol{\mu}[k])^\top({}^s\boldsymbol{\pi}_{({}^i\boldsymbol{\mu})}[k+1]) \nonumber \\
& + \sum_{k=0}^{K-1}(\boldsymbol{h} - \boldsymbol{H}_{\mu}\boldsymbol{\mu}[k] - \boldsymbol{H}_{\nu}{}^s\widehat{\boldsymbol{\nu}}[k])^\top({}^s\boldsymbol{\lambda}_{({}^i\boldsymbol{\mu})}[k]) \nonumber \\
&\overset{\ref{property:a} \ref{property:b} \ref{property:c}}{=} \boldsymbol{x}_0^\top(\boldsymbol{\pi}_{({}^i\boldsymbol{\mu})}[1]) + \sum_{k=0}^{K-2}(\boldsymbol{G}_{\mu}\boldsymbol{\mu}[k])^\top(\boldsymbol{\pi}_{({}^i\boldsymbol{\mu})}[k+2]) \nonumber \\
& + \sum_{k=0}^{K-2}(\boldsymbol{h} - \boldsymbol{H}_{\mu}\boldsymbol{\mu}[k] - \boldsymbol{H}_{\nu}\widehat{\boldsymbol{\nu}} [k+1])^\top(\boldsymbol{\lambda}_{({}^i\boldsymbol{\mu})}[k+1]) \nonumber \\
&\overset{\ref{property:d} \ref{property:e}}{=} \boldsymbol{x}_0^\top(\boldsymbol{\pi}_{({}^i\boldsymbol{\mu})}[1]) + \sum_{k=0}^{K-1}(\boldsymbol{G}_{\mu}{}^i\boldsymbol{\mu}[k])^\top(\boldsymbol{\pi}_{({}^i\boldsymbol{\mu})}[k+1]) \nonumber \\
& + \sum_{k=0}^{K-1}(\boldsymbol{h} - \boldsymbol{H}_{\mu}{}^i\boldsymbol{\mu}[k] - \boldsymbol{H}_{\nu}\widehat{\boldsymbol{\nu}}[k])^\top(\boldsymbol{\lambda}_{({}^i\boldsymbol{\mu})}[k]) \nonumber \\
& \overset{\ref{property:f}}{=} \boldsymbol{x}_0^\top(\boldsymbol{\pi}_{({}^i\boldsymbol{\mu})}[0]) + \sum_{k=0}^{K-1}(\boldsymbol{G}_{\mu}{}^i\boldsymbol{\mu}[k])^\top(\boldsymbol{\pi}_{({}^i\boldsymbol{\mu})}[k+1]) \nonumber \\
& + \sum_{k=0}^{K-1}(\boldsymbol{h} - \boldsymbol{H}_{\mu}{}^i\boldsymbol{\mu}[k] - \boldsymbol{H}_{\nu}\widehat{\boldsymbol{\nu}}[k])^\top(\boldsymbol{\lambda}_{({}^i\boldsymbol{\mu})}[k]) < 0 \nonumber
\end{align}


The final inequality follows from condition \eqref{eq:farkas_d} with the original certificates $\boldsymbol{\lambda}_{({}^i\boldsymbol{\mu})}$ and $\boldsymbol{\pi}_{({}^i\boldsymbol{\mu})}$ for the problem $\mathcal{P}(\boldsymbol{x}_0, {}^i\boldsymbol{\mu}, \widehat{\boldsymbol{\nu}})$, which is infeasible by construction.
\end{proof}

Note that in our proof, each element of $\boldsymbol{\nu}[k]$ indicates if a specific PoI is visited at time $k$ through one-hot encoding for our bipedal TAMP problem. This theorem instead adopts binary encoding, where $\boldsymbol{\nu}[k] \in \{0,1\}^{n_\nu}$ can represent up to $2^{n_\nu}$ distinct PoIs. The conversion between the two encodings is straightforward—for example, PoI 3 corresponds to the binary vector $[0,0,1]$.

The time-shifted cuts can be generated iteratively: starting from the infeasible schedule $\widehat{\boldsymbol{\nu}}$, we continue shifting (generating cuts for $s = 1, 2, ...$) until condition 2 becomes invalid, i.e., when $\boldsymbol{C}\boldsymbol{x}_0 + \boldsymbol{D}\boldsymbol{u}[0] + \boldsymbol{H}_{\nu}\widehat{\boldsymbol{\nu}}[s] \leq \boldsymbol{h}$ becomes infeasible. This determines the maximum allowable shift $s_{\max}$.

While Theorem \ref{theorem2} assumes the BSP is formulated as MIQP, our specific BSP formulation is formulated as NLPs and solved with IPOPT to local optimality for computational efficiency. However, the nonlinear collision avoidance constraint \eqref{eqn:softmin_collision} and the nonlinear kinematic constraint \eqref{eqn:kinematics} can equivalently be formulated into MIQP constraints as in \cite{deits2014footstep} and solved with an MIQP solver. 

Condition 1 holds in many common scenarios where the robot starts from an equilibrium stance. In contrast, condition 3 is more restrictive and may not apply to all problems. Nevertheless, time-shifted cuts can still be applied in such cases. The potential issue is that they might eliminate feasible solutions, causing a problem that was originally feasible to appear infeasible. If this happens, reducing the maximum shift amount $s_{\max}$ can help recover feasible solutions. The key requirement of Theorem \ref{theorem2} is to establish that $\boldsymbol{\lambda}_{({}^i\boldsymbol{\mu})}[0] = \boldsymbol{0}$; condition 3 provides one sufficient—but not necessary—justification, and alternative justifications may exist for specific problems. Nevertheless, while the theorem's completeness guarantees do not strictly apply to our NLP formulation (and many other problems), the time-shifted cuts remain to be effective heuristics that significantly reduce the number of BD iterations.

Our time-shifted cuts can be viewed as a variant of local branching for BD \cite{rei2009accelerating}, where we leverage subproblem structure to identify multiple theoretically guaranteed feasibility cuts without additional searching in the neighborhood. When the theorem's conditions hold, this technique can provide strong cuts ``for free'', making it a valuable unless cut management becomes a concern.



\section{ADMM Formulation for STL-Based TAMP of Bipedal Locomotion}
\label{formulation_ADMM}
In this appendix, we provide the detailed formulation of the ADMM method used as a baseline in our experiments. The ADMM framework decomposes the bipedal TAMP problem into two subproblems that are solved iteratively: an MICP that handles discrete task scheduling under STL specifications with simplified kinematic constraints, and an NLP that enforces full kinematic feasibility and obstacle avoidance. Unlike the BD approach presented in the main text, both subproblems in ADMM maintain separate copies of the decision variables and rely on consensus penalties to coordinate their solutions.

The MICP subproblem includes the LIP dynamics constraints \eqref{eqn:LIP}, relaxed kinematic reachability constraints \eqref{eqn:circular_reach}, control input bounds \eqref{eqn:velocity_bound}, STL predicate and logic constraints \eqref{eqn:bigm}, \eqref{eqn:connectives}, and the region assignment constraint \eqref{eqn: oneregion}. On the other hand, the NLP subproblem incorporates the same LIP dynamics \eqref{eqn:LIP}, the exact kinematics constraints \eqref{eqn:kinematics}, \eqref{eqn:stability}, obstacle avoidance \eqref{eqn:softmin_collision}, and STL satisfaction \eqref{eqn:connectives}. Let $f^{\text{MIP}}$ and $f^{\text{NLP}}$  denote the objective functions of the MICP and NLP subproblems, respectively (excluding the consensus cost), and let $f^{\text{css}}$ represent the consensus penalty. The MICP subproblem in the ADMM algorithm is formulated as follows: 
\begin{equation}
\label{eq:admm_micp}
\begin{aligned}
\underset{\mathbf{X}, \mathbf{U}, \mathbf{Z} \in \mathbb{B}^{|\mathbf{Z}|}}{\text{minimize}} \quad & f^{\text{MIP}}+f^{\text{css}} \\
\text{s.t.} \quad & 
\begin{cases}
\eqref{eqn:bigm}, \ \forall k\in \mathcal{K} \ \forall \xi \in \Xi; \\ 
\eqref{eqn:connectives}; \\ 
\eqref{eqn: oneregion}, \ \forall k \in \mathcal{K}; \\
\eqref{eqn:LIP}\ \eqref{eqn:circular_reach}\ \eqref{eqn:velocity_bound}, \ \forall k \in \mathcal{K} \setminus \{K\}
\end{cases}
\end{aligned}
\end{equation}
and the NLP subproblem is:
\begin{equation}
\label{eq:admm_nlp}
\begin{aligned}
&\underset{\mathbf{X}, \mathbf{U}, \mathbf{Z} \in [0, 1]^{|\mathbf{Z}|}}{\text{minimize}} \quad f^{\text{NLP}} + f^{\text{css}}\\
&\text{s.t.}
\quad\quad
\begin{cases}
\eqref{eqn:LIP}\ \eqref{eqn:kinematics}\ \eqref{eqn:stability}, \ \forall k \in \mathcal{K}\setminus\{K\};\\
\eqref{eqn:softmin_collision}, \ \forall \lambda \in \Xi_{\text{obs}}, \ \forall k \in \mathcal{K}; \\
\eqref{eqn:connectives};
\end{cases}
\end{aligned}
\end{equation}

In our BD decomposition method, the cost is separated such that the BMP (MICP) minimizes  $f^{\text{time}}$, which encourages earlier task completion, while the BSP (NLP) minimizes $f^{\text{loco}}$, which penalizes undesirable bipedal walking behaviors. In contrast, the standard ADMM decomposition typically assigns all original cost terms along with consensus penalties to one subproblem, while the other subproblem minimizes only the consensus terms. This setup effectively treats the second subproblem as a projection step, refining the solution to satisfy constraints not enforced in the first subproblem. 
In our bipedal TAMP problem, however, directly incorporating $f^{\text{loco}}$ into the MICP subproblem is infeasible due to its nonlinear components, such as $\boldsymbol{R}(\theta_k)^\top \boldsymbol{u}_k^{\text{ft}}$ and $\boldsymbol{R}(\theta_k)^\top \boldsymbol{v}_k$, which would make the problem a mixed-integer nonlinear program. To address this challenge, we propose two variants of ADMM formulations as below.

\begin{algorithm}[t!]
\caption{ADMM for Bipedal TAMP}
\label{algorithm_admm}
\textbf{Input:} Problem formulation, weights $w_v$, parameter $\rho$\\
\textbf{Initialization:} Set $i = 0$ and $\lambda_v = 0, \forall v \in \mathcal{V}$\\
\While{not converged}{
    \textbf{Step 1:} Solve MICP subproblem \eqref{eq:admm_micp} with consensus to $v^{\text{NLP}}_{i-1}$, obtain $v^{\text{MIP}}_i$\\
    
    \textbf{Step 2:} Solve NLP subproblem \eqref{eq:admm_nlp} with consensus to $v^{\text{MIP}}_i$, obtain $v^{\text{NLP}}_i$\\
    
    \textbf{Step 3:} Update dual variables\\
    \quad $\lambda_v \leftarrow \lambda_v + (v^{\text{MIP}}_i - v^{\text{NLP}}_i)$ for all $v \in \mathcal{V}$\\
    
    \textbf{Step 4:} Update penalty weights\\
    \quad $w_v \leftarrow w_v/\rho, \forall v \in \mathcal{V}$\\
    
    $i \leftarrow i + 1$
}
\textbf{Return:} Feasible solution $(v^{\text{MIP}}_i, v^{\text{NLP}}_i)$
\end{algorithm}

The first ADMM variant assigns $f^{\text{loco}}$ to the objective function of the NLP subproblem. In this case, we define $f^{\text{MIP}} = f^{\text{time}}$, which captures only the task completion cost, and $f^{\text{NLP}} = \sum_{k=1}^{K} f_k^{\text{loco}}$, which accounts solely for the locomotion-related cost. Let $\mathcal{V}$ denote the set of all decision variables shared between the two subproblems, where $\mathcal{V} = \{\mathbf{X}, \mathbf{U}, \mathbf{Z}\}$ contains all the state, control, and binary variables for this first ADMM variant. For each $v \in \mathcal{V}$, we use $v^{\text{MIP}}$ to represent its copy in the MICP subproblem and $v^{\text{NLP}}$ for its counterpart in the NLP subproblem. Additionally, we let $\lambda_v$ represent the dual variable associated with each $v$, and $w_v$ stands for the corresponding penalty weight. The consensus cost takes the form of:
\begin{align*}
f^{\text{css}} = \sum_{v \in \mathcal{V}} w_v \|v^{\text{MIP}} - v^{\text{NLP}} + \lambda_v\|^2
\end{align*}

For the second ADMM variant, we introduce additional consensus variables $s_k=\sin(\theta_k)$, $c_k=\cos(\theta_k)$, $s_{x,k} = s_k \cdot x_k$, $s_{y,k} = s_k\cdot y_k$, $c_{x,k} = c_k\cdot x_k$, $c_{y,k} = c_k\cdot y_k$, $s_{\dot{x},k} = s_k\cdot \dot{x}_k$, $s_{\dot{y},k} = s_k\cdot \dot{y}_k$, $c_{\dot{x},k} = c_k\cdot \dot{x}_k$, and $c_{\dot{y},k} = c_k\cdot \dot{y}_k$ for the bilinear products. These auxiliary variables allow us to express $f^{\text{loco}}$ as a quadratic function of the new variables, to fit in the MICP objective function. In this setting, $f^{\text{MIP}} = f^{\text{time}} + \sum_{k=1}^{K} f_k^{\text{loco}}$ and $f^{\text{NLP}} = 0$. The set of consensus variables is expanded to $\mathcal{V} = \{\mathbf{X}, \mathbf{U}, \mathbf{Z}\} \cup \{s_k, c_k, s_{x,k}, s_{y,k}, c_{x,k}, c_{y,k}, s_{\dot{x},k}, s_{\dot{y},k}, c_{\dot{x},k}, c_{\dot{y},k} : \forall k \in \mathcal K\}$. We add additional equality constraints in the NLP subproblem to enforce the nonlinear relationships, such as $s_{x,k} = s_k \cdot x_k$.

As such, the constraints in ADMM's MICP subproblem are identical to those in the BMP, with the only difference lying in the objective function. There are more substantial differences between ADMM's NLP subproblem and the BSP. Specifically, the BSP does not include the binary variable $\mathbf{Z}$, as it is fixed by the BMP. However, in ADMM, the NLP subproblem retains a relaxed version of $\mathbf{Z}$, allowing it to take continuous values in the range of $[0, 1]^{|\mathbf{Z}|}$, and relies on the consensus cost to drive these values toward binary solutions. 
Furthermore, unlike BD, ADMM’s NLP is not decomposed into multiple trajectory segments for parallel computation, since $\mathbf{Z}$ remains a variable in the optimization that prevents such segmentation. Similar to BD, the NLP subproblem in ADMM accounts for less than 10\% of the total computation time, making decomposition unnecessary from a computation performance standpoint.

The ADMM algorithm proceeds iteratively, alternating between the two subproblems. At $i^{\rm th}$ iteration, we first solve the MICP subproblem to obtain $v^{\text{MIP}}_i$ with consensus penalties relative to $v^{\text{NLP}}_{i-1}$ (except for the initial iteration $i=0$, where no consensus terms are applied). Next, we solve the NLP subproblem to obtain $v^{\text{NLP}}_i$ with alignment to $v^{\text{MIP}}_{i}$. The dual variables are then updated according to $\lambda_{v,i+1} = \lambda_{v,i} + (v^{\text{MIP}}_i - v^{\text{NLP}}_i)$, and the penalty weights are updated by $w_v \leftarrow w_v/\rho$ with parameter $\rho > 1$ to improve convergence. The algorithm continues until the residual $\|v^{\text{MIP}}_i - v^{\text{NLP}}_i\|_\infty$ falls below a predefined threshold. The complete procedure is summarized in Algorithm 2.

In our setting, we set $w_v$ based on variable types: $10^4$ for positions, $10^5$ for velocities, angles, and angular velocities, $10^3$ for fooothold locations, and $10^2$ for binary variables, with $\rho$ selected bewteen $1.0$ and $3.0$. We test both approaches and find similar performance in terms of convergence iterations, computation time, and solution quality.

\end{appendices}

\bibliographystyle{IEEEtran}
\bibliography{references.bib}

\end{document}